\documentclass{article}

\usepackage[nonanonymous]{neurips_2026}

\usepackage[utf8]{inputenc}
\usepackage[T1]{fontenc}
\usepackage{hyperref}
\usepackage{url}
\usepackage{xurl}
\usepackage{booktabs}
\usepackage{amsfonts}
\usepackage{amsmath}
\usepackage{amssymb}
\usepackage{mathtools}
\usepackage{amsthm}
\usepackage{nicefrac}
\usepackage{microtype}
\usepackage{xcolor}
\usepackage{textcomp}
\usepackage{graphicx}
\usepackage{wrapfig}
\usepackage[section]{placeins}
\usepackage{subcaption}
\usepackage{caption}
\usepackage[inline]{enumitem}
\usepackage{listings}
\usepackage{siunitx}
\usepackage{float}
\usepackage{longtable}
\usepackage{multirow}
\usepackage{threeparttable}
\usepackage{pdflscape}
\usepackage{array}
\usepackage{makecell}
\usepackage{tikz}
\usepackage[capitalize,noabbrev]{cleveref}
\usepackage{additional_macros}
\hypersetup{hidelinks}

\makeatletter
\renewcommand{\@maketitle}{%
  \vbox{%
    \hsize\textwidth
    \linewidth\hsize
    \vskip 0.04in
    \hrule height 4\p@
    \vskip 0.12in
    \vskip -\parskip
    \centering
    {\LARGE\bf \@title\par}
    \vskip 0.18in
    \vskip -\parskip
    \hrule height 1\p@
    \vskip 0.08in
    \begin{tabular}[t]{c}\rule{\z@}{14\p@}\@author\end{tabular}%
    \vskip 0.11in \@minus 0.05in
  }
}
\makeatother
\renewenvironment{abstract}%
{\vskip 0.05in\centerline{\large\bf Abstract}\vspace{0.4ex}\begin{quote}}%
{\par\end{quote}\vskip 0.5ex}
\long\def\skillsbenchAuthorNamesBlock{%
  \begingroup\normalfont
  \makebox[0pt][c]{%
    \begin{minipage}{\textwidth}
    \centering\setlength{\parskip}{0pt}%
    {\bfseries \benchAuthor{Xiangyi Li}{1,*}, \benchAuthor{Yimin Liu}{1,2,*},
    \benchAuthor{Wenbo Chen}{3,*,\textdagger}, \benchAuthor{Bingran You}{1,4,*},
    \benchAuthor{Zonglin Di}{5,\textdaggerdbl},\\[-0.05em]
    \benchAuthor{Yifeng He}{6,\textdaggerdbl},
    \benchAuthor{Shenghan Zheng}{7,\textdaggerdbl},
    \benchAuthor{Kyoung Whan Choe}{8,\textdaggerdbl},
    \benchAuthor{Jiankai Sun}{9,\textdaggerdbl},
    \benchAuthor{Shuyi Wang}{9,\textdaggerdbl}, \benchAuthor{Chujun Tao}{14,\textdaggerdbl},
    \benchAuthor{Binxu Li}{10,\textdaggerdbl}, \benchAuthor{Xuandong Zhao}{4,\textdaggerdbl},
    \benchAuthor{Hejia Geng}{11}, \benchAuthor{Xiaojun Wu}{35},
    \benchAuthor{Junwei Zhou}{9}, \benchAuthor{Xiaokun Chen}{12},
    \benchAuthor{Hanwen Xing}{13}, \benchAuthor{Yubo Li}{14},
    \benchAuthor{Qunhong Zeng}{9}, \benchAuthor{Di Wang}{15},
    \benchAuthor{Yuanli Wang}{16}, \benchAuthor{Roey Ben Chaim}{17},
    \benchAuthor{Penghao Jiang}{18}, \benchAuthor{Haotian Shen}{9},
    \benchAuthor{Luyang Kong}{9}, \benchAuthor{Xinyi Liu}{9},
    \benchAuthor{Runhui Wang}{9}, \benchAuthor{Xuanqing Liu}{9},
    \benchAuthor{Jiachen Li}{19}, \benchAuthor{Xin Lan}{20},
    \benchAuthor{Yueqian Lin}{21}, \benchAuthor{Wengao Ye}{11},
    \benchAuthor{Junwei He}{22}, \benchAuthor{Songlin Li}{12},
    \benchAuthor{Yue Zhang}{23}, \benchAuthor{Yipeng Gao}{13},
    \benchAuthor{Yijiang Li}{24}, \benchAuthor{Ze Ma}{25},
    \benchAuthor{Liqiang Jing}{23}, \benchAuthor{Tianyu Wang}{9},
    \benchAuthor{Kaixin Li}{9}, \benchAuthor{Yiqi Xue}{13},
    \benchAuthor{Haoran Lyu}{9}, \benchAuthor{Yizhuo He}{14},
    \benchAuthor{Yuchen Tian}{9}, \benchAuthor{Shutong Wu}{26},
    \benchAuthor{Bowei Wang}{9}, \benchAuthor{Yixuan Gao}{27},
    \benchAuthor{Bo Chen}{9}, \benchAuthor{Litong Liu}{28},
    \benchAuthor{Sikai Cheng}{28}, \benchAuthor{Jiajun Bao}{14},
    \benchAuthor{Shuaicheng Tong}{28}, \benchAuthor{Shuwen Xu}{9},
    \benchAuthor{Terry Yue Zhuo}{9}, \benchAuthor{Tinghan Ye}{28},
    \benchAuthor{Qi Qi}{9}, \benchAuthor{Miao Li}{28},
    \benchAuthor{Longtai Liao}{9}, \benchAuthor{Zelin Tan}{34},
    \benchAuthor{Chang Shi}{19},
    \benchAuthor{Xilin Tang}{29}, \benchAuthor{Srinath Tankasala}{3,\textdagger},
    \benchAuthor{Boqin Yuan}{24}, \benchAuthor{Yaoyao Qian}{30},
    \benchAuthor{Jianhong Tu}{5}, \benchAuthor{Chenguang Wang}{5},
    \benchAuthor{Yizhou Sun}{31},\\[-0.05em]
    \benchAuthor{Wei Wang}{31},
    \benchAuthor{Aaron Taylor}{33},
    \benchAuthor{Ziyue Yang}{6}, \benchAuthor{Changkun Guan}{28},
    \benchAuthor{Zhikang Dong}{32},\\[-0.05em]
    \benchAuthor{Xinyu Zhang}{36}, \benchAuthor{Steven Dillmann}{12},
    \benchAuthor{Han-chung Lee}{9},
    \benchAuthor{Dawn Song}{4}\par}
    \end{minipage}%
  }%
  \endgroup
}
\newcommand{\skillsbenchAffilFootnote}{%
  {\renewcommand{\thefootnote}{}%
  \footnotetext{\scriptsize \textsuperscript{1}BenchFlow, \textsuperscript{2}OSU, \textsuperscript{3}Amazon, \textsuperscript{4}UC Berkeley, \textsuperscript{5}UC Santa Cruz, \textsuperscript{6}UC Davis, \textsuperscript{7}Dartmouth, \textsuperscript{8}RLWRLD, \textsuperscript{9}Independent, \textsuperscript{10}Princeton University, \textsuperscript{11}Oxford University, \textsuperscript{12}Stanford University, \textsuperscript{13}USC, \textsuperscript{14}CMU, \textsuperscript{15}Foxconn, \textsuperscript{16}BU, \textsuperscript{17}Zenity, \textsuperscript{18}UNSW, \textsuperscript{19}UT Austin, \textsuperscript{20}MSU, \textsuperscript{21}Duke University, \textsuperscript{22}ByteDance, \textsuperscript{23}UT Dallas, \textsuperscript{24}UC San Diego, \textsuperscript{25}Columbia University, \textsuperscript{26}University of Rochester, \textsuperscript{27}Cornell Tech, \textsuperscript{28}Georgia Tech, \textsuperscript{29}Cornell University, \textsuperscript{30}NEU, \textsuperscript{31}UCLA, \textsuperscript{32}Snap Inc., \textsuperscript{33}Fanshawe College, \textsuperscript{34}University of Science and Technology of China, \textsuperscript{35}HKUST(GZ), \textsuperscript{36}Anyscale\\[0.25em] \textsuperscript{*}Equal contribution. \textsuperscript{\textdaggerdbl}Core contribution. \textsuperscript{\textdagger}This work was conducted outside the author's role at Amazon. Correspondence: Xiangyi Li at \href{mailto:xiangyi@benchflow.ai}{xiangyi@benchflow.ai}.}}%
}
\newcommand{\FigOneBlock}{%
  \begingroup
  \captionsetup[figure]{font=small,skip=5pt}%
  \noindent\begin{minipage}{\textwidth}
  \centering
  \includegraphics[width=\textwidth]{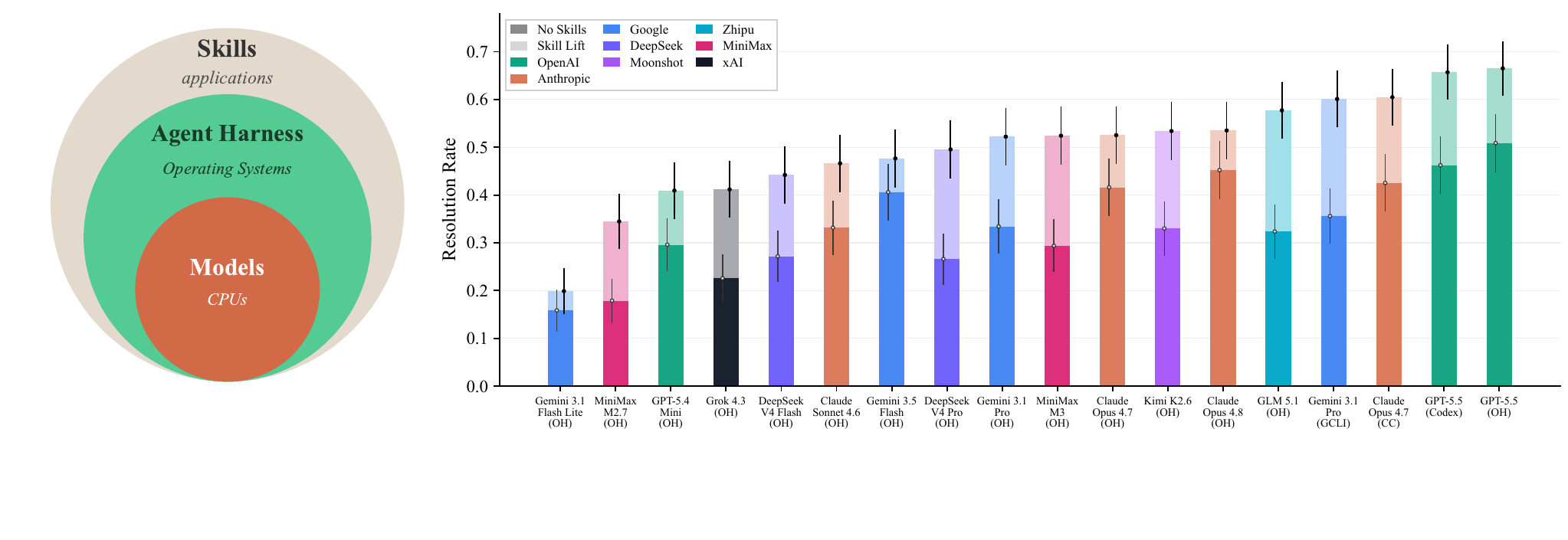}\par
  \captionof{figure}{Agent architecture stack and resolution rates across 18 model--harness configurations on 87 SkillsBench tasks. Each bar stacks the no-Skills baseline and curated-Skills lift; open/filled interval markers denote 95\% CIs for the baseline/total. Bars are ordered by Curated-Skills pass rate and colored by model family. Harness abbreviations: OH = OpenHands, CC = Claude Code, GCLI = Gemini CLI.}\label{fig:skills_results}%
  \end{minipage}%
  \endgroup
}

\makeatletter
\renewcommand{\@notice}{}
\makeatother

\definecolor{posgreen}{rgb}{0.13,0.54,0.13}
\definecolor{negred}{rgb}{0.80,0.20,0.15}
\newcommand{\ppos}[1]{\textcolor{posgreen}{+#1}}
\newcommand{\nneg}[1]{\textcolor{negred}{--#1}}

\renewcommand{\arraystretch}{1.35}
\theoremstyle{plain}

\theoremstyle{definition}

\theoremstyle{remark}

\newcommand{\benchmarkName}{\textsc{SkillsBench}}
\newcommand{\benchAuthor}[2]{\mbox{#1\textsuperscript{#2}}}

\title{\raisebox{-0.029in}{\includegraphics[height=0.23in]{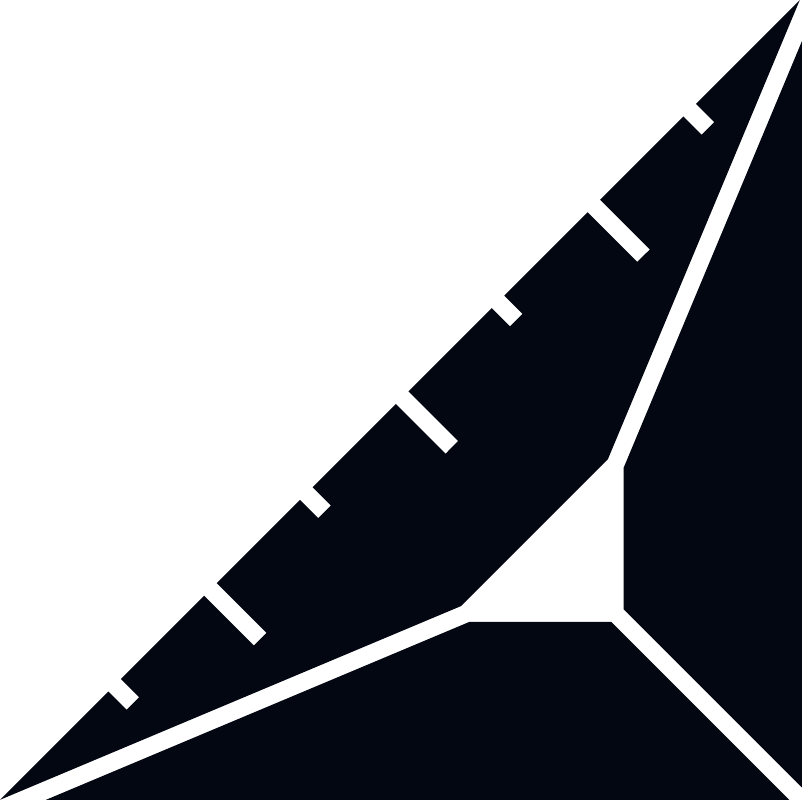}}\hspace{6pt}SkillsBench: Benchmarking How Well Agent Skills Work Across Diverse Tasks}

\author{\skillsbenchAuthorNamesBlock}

\begin{document}
\nolinenumbers

\maketitle
\skillsbenchAffilFootnote

\begin{abstract}
Agent Skills are structured packages of procedural knowledge that augment large language model (LLM) agents at inference time. Despite rapid adoption, there is no standard way to measure whether they actually help. We present \benchmarkName{}, a benchmark whose current inventory contains 87 tasks across 8 domains paired with curated Skills and deterministic verifiers. Our latest aggregate evaluation runs the 87-task benchmark under matched no-Skills and curated-Skills conditions for 18 model--harness configurations. Curated Skills raise the average pass rate from 33.9\% to 50.5\% (+16.6 percentage points; 25.5\% normalized gain), with configuration-level gains ranging from +4.1 to +25.7\,pp. Focused Skills with at most three modules outperform larger or exhaustive bundles, and smaller models with Skills can match larger models without them. \benchmarkName{} establishes paired evaluation as the foundation for rigorous measurement of Skill efficacy on agentic, expertise-heavy work.
\end{abstract}

\vspace{0.3em}\FigOneBlock\par\vspace{0.6em}

\section{Introduction}
\label{sec:intro}

AI agents are now deployed in production workflows, from software engineering~\citep{anthropic2025claudecode,google2025geminicli,openai2025codexcli}, where they sustain task horizons of up to 10 hours of human effort, to expertise-heavy domains beyond it. A fundamental tension follows: foundation models provide broad capabilities but lack the procedural knowledge a specialist brings on day one, while fine-tuning each domain is expensive and sacrifices generality~\citep{brown2020languagemodelsfewshotlearners, ouyang2022training, yao2023reactsynergizingreasoningacting}.

\textbf{Agent Skills}~\citep{anthropic2025agentskills} are an emerging solution: structured packages of instructions, code templates, resources, and reference material that augment agents at inference time without modifying model weights. Skills encode standard operating procedures, domain conventions, and task heuristics as modular artifacts mediated by the agent harness~\citep{sutton1999between, sumers2024cognitive}.

Community Skills ecosystems have already grown to 2{,}014{,}000 source-partitioned Skills in our construction snapshot (Figure~\ref{fig:pipeline}; Appendix~\ref{app:ecosystem}). Yet despite this proliferation, no benchmark systematically asks: \textbf{how much do Skills actually help, and when do they fail?} Existing agent benchmarks~\citep{liu2023agentbench, merrill2026terminalbenchbenchmarkingagentshard, jimenez2024swebench, zhou2024webarena, xie2024osworld, koh2024visualwebarena, trivedi2024appworld, yang2023intercode, chan2025mlebench, zhuo2025bigcodebench} measure raw capability in isolation, asking how well a model performs task $X$ while folding model, harness, and augmentation effects into one pass rate. They do not answer the deployment question: \emph{will adding this Skill help my agent on this task, and by how much?} \autoref{fig:skills_results} illustrates the layered architecture and previews our main result: curated Skills improve resolution rates across 18 model--harness configurations by +16.6\,pp on average.

We introduce \benchmarkName{}, the first benchmark that treats Skills as a first-class evaluation artifact. Our contribution is two-tier:

\begin{itemize}[nosep,leftmargin=*]
\item \textbf{Narrow contribution: a quantitative answer to ``how much do Skills help?''} On the 87-task benchmark, evaluated under matched no-Skills and curated-Skills conditions across 18 model--harness configurations, curated Skills lift task-macro pass rate from 33.9\% to 50.5\% (+16.6\,pp; 25.5\% normalized gain), with substantial configuration-level heterogeneity (+4.1 to +25.7\,pp).

\item \textbf{Broad contribution: a paired-evaluation framework for agent augmentation, generalizable beyond Skills.} The same paired (with vs.\ without) protocol, contributor-driven sourcing, leakage-controlled task-to-artifact decoupling, and BenchFlow~\citep{benchflowteam2026benchflow} containerized harness can evaluate other artifacts (retrieval pipelines, memory stores, scaffolding) without confounding model and augmentation effects. We open-source the benchmark, harness, and public trajectories/results so practitioners can test their own Skill libraries before shipping.
\end{itemize}

\section{Background}
\label{sec:Back}




\paragraph{Skill Definition}
In this paper, we define a \textbf{Skill} as a reusable, file-system-based procedural package for a class of agent tasks. 
Each Skill contains a required \texttt{SKILL.md} file with natural-language instructions and may optionally include auxiliary resources such as scripts, templates, reference files, or worked examples.

\paragraph{Skill Augmentation}
We use \textbf{Skill augmentation} to refer to the inference-time mechanism by which an agent harness makes relevant Skills available to an agent for solving a task. 
Compared with other runtime augmentation paradigms, Skill augmentation is \emph{modular and reusable}, provides \emph{procedural guidance} rather than factual context alone, can include \emph{executable resources}, and is \emph{cross-model portable} because Skills are represented as files rather than model parameters (\autoref{tab:comparison}).

\begin{table}[!t]
\centering
\small
\caption{Comparison of runtime augmentation paradigms. Skills combine modular packaging, procedural guidance, executable resources, and portability.}
\label{tab:comparison}
\vspace{0.45em}
\begin{tabular}{lcccc}
\toprule
& \textbf{Prompts} & \textbf{RAG} & \textbf{Tools} & \textbf{Skills} \\
\midrule
Modular/reusable & \texttimes & \checkmark & \checkmark & \checkmark \\
Procedural guidance & Limited & \texttimes & \texttimes & \checkmark \\
Executable resources & \texttimes & \texttimes & \checkmark & \checkmark \\
Cross-model portable & \checkmark & \checkmark & \checkmark & \checkmark \\
\bottomrule
\end{tabular}
\end{table}

\paragraph{Agent Harness}
An \textbf{agent harness} is the execution layer that wraps an LLM and connects it to its environment~\citep{harness-engineering-codex,leehanchung2026the-training-grounds}. 
It exposes tools, manages files and workspace state, executes scripts, returns observations, and, in Skill-augmented agents, discovers and loads relevant Skills during inference. 

Thus, a Skill is neither model weights nor an executable tool by itself; it becomes actionable through a compatible harness that injects its instructions and exposes its resources.


\section{\benchmarkName{}}
\label{sec:SkillsBench}

A benchmark that measures Skill efficacy is only as credible as its filtering pipeline: if a Skill can encode task-specific answers, the measurement collapses into instruction-following. \benchmarkName{} therefore treats construction as part of the contribution. We define valid tasks, source candidates from a vetted contributor community, and filter them through automated gates and human review. The current evaluated inventory contains 87 tasks across 8 domains, drawn from 400 submissions by 142 contributors. Each task is a containerized unit with fixed data, an oracle solution, and a deterministic verifier, evaluated under matched no-Skills and Skills-augmented conditions to isolate the Skill's contribution from the model--harness configuration.

\textbf{Skills as expertise, not answers.} \benchmarkName{} requires Skills to provide domain expertise for a \emph{class} of problems, never the solution to a specific instance. We enforce this in two ways. First, contributors author Skills independently of the benchmark---from public repositories or prior domain experience---so the experiment measures use of pre-existing expertise. Second, task instructions never name which Skills to use; agents discover and activate Skills through the standard progressive-disclosure mechanism~\citep{anthropic2025agentskills}. Without these constraints, a Skill becomes a hidden answer key.

\begin{figure}[!t]
  \centering
  \includegraphics[width=0.9\textwidth]{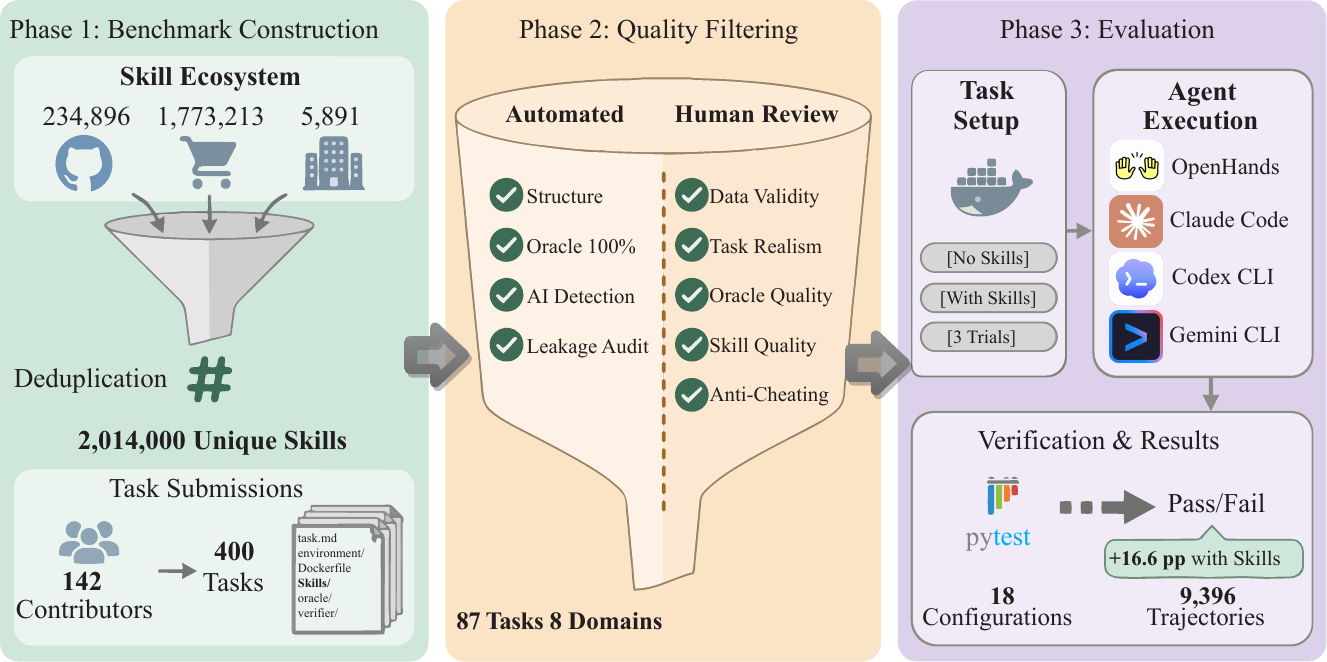}
  \caption{\textbf{\benchmarkName{} pipeline overview.} Construction aggregates 2{,}014{,}000 source-partitioned Skills and 400 candidate task submissions from 142 contributors. Filtering applies automated checks and human review, retaining an 87-task public evaluation aggregate across 8 domains. Evaluation runs each task under matched Skill-access conditions across 18 model--harness configurations on BenchFlow~\citep{benchflowteam2026benchflow}, producing 9{,}396 selected public result files.}
  \label{fig:pipeline}
\end{figure}

\textbf{Task principles.} Beyond the Skill boundary, every task must be \textit{authentic} real work, \textit{verifiable} by deterministic pass/fail tests rather than LLM-as-a-judge~\citep{wang2023execution, brown_verifiers_2025}, \textit{difficult} because of the problem rather than artificial instruction confusion, and \textit{solvable} end-to-end by an oracle agent without pre-baked answers or hard-coded magic numbers. Instructions are human-authored; if LLMs help refine wording, a human owns the final iteration. We reject classroom-style tasks, made-up scenarios, toy datasets, and purely synthetic data.

\textbf{Sourcing and filtering.} We grew a 1{,}400-member contributor community, onboarded 200+ task authors through scoping interviews, and ingested 400 candidate submissions from the public PR history. Every PR clears four automated gates before human review: structural integrity, oracle execution, instruction provenance (AI-text detector plus human label, yielding a 100\% human-authored release set), and Skill-to-solution leakage (rejecting task-specific filenames, paths, magic numbers, verbatim oracle commands, or expected outputs). PRs that pass receive at least one 30-minute maintainer review; tasks with no measurable separation between conditions are rejected as low-signal. The current public evaluation aggregate retains 87 tasks; per-gate rejection counts and the full rubric are in Appendix~\ref{app:skillsbench-construction-details}.

\begin{figure}[!t]
  \centering
  \captionsetup{skip=2pt}
  \includegraphics[width=\textwidth]{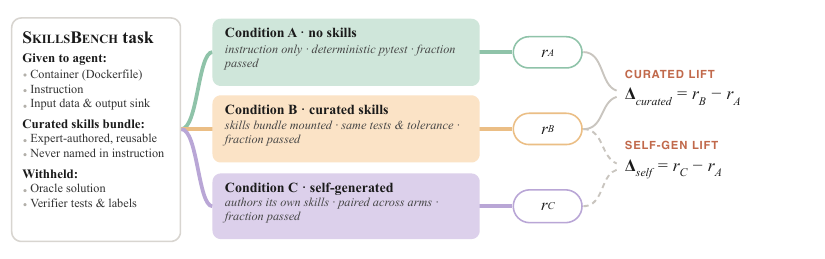}
  \caption{\textbf{Anatomy of a \benchmarkName{} task.} Each task ships with a containerized environment, human-authored instruction, expert-curated Skills bundle, and withheld oracle/verifier/labels. The arms differ only in Skill access (A: instruction alone; B: curated bundle mounted for the agent to discover; C: the agent authors its own skill documents, then solves with them); one deterministic \texttt{pytest} battery scores every arm as the fraction of checks passed, $r_A,r_B,r_C\in[0,1]$. Skill efficacy: $\Delta_{\text{curated}}=r_B-r_A$, and $\Delta_{\text{self}}=r_C-r_A$ against the same baseline.}
  \label{fig:task-anatomy}
\end{figure}

\textbf{Task specification.} A \benchmarkName{} task is a self-contained module with four components (Figure~\ref{fig:task-anatomy}), following the Agent Skills format~\citep{anthropic2025agentskills}: a human-authored \emph{instruction}, a Docker \emph{environment} with task data and a \texttt{skills/} subdirectory, an \emph{oracle} reference solution that must pass the verifier, and a deterministic \texttt{pytest} \emph{verifier}. Each Skill folder uses the standard layout: a required \texttt{SKILL.md} plus optional scripts, references, and assets.

\begin{figure}[!b]
  \centering
  \includegraphics[width=\linewidth]{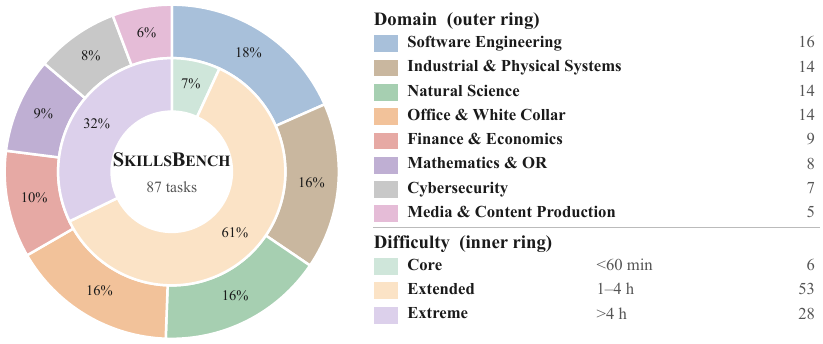}
  \caption{\benchmarkName{} spans 8 domains (87 tasks total), weighted toward production workflows rather than classroom problem sets. The inner ring shows the difficulty stratification; domain definitions and the per-task mapping appear in \autoref{tab:task-domain-mapping}.}
  \label{fig:category_distribuion}
\end{figure}

\textbf{Composition.} The current task inventory contains 87 tasks across 8 domains (Figure~\ref{fig:category_distribuion}), stratified by estimated human-specialist completion time without AI assistance: 6 Core ($<$60\,min), 53 Extended (1--4\,h), 28 Extreme ($>$4\,h). Per-domain counts range from $N=5$ to $N=16$.

\section{Experiment}
\label{sec:experimentation}

\textbf{Models, harnesses, and conditions.}
For the latest aggregate in \autoref{tab:main-results}, we evaluate 18 configurations across four terminal-agent harnesses: OpenHands, Gemini CLI~\citep{google2025geminicli}, Claude Code~\citep{anthropic2025claudecode}, and Codex CLI~\citep{openai2025codexcli}. Each task runs under two matched conditions: \emph{no Skills} (the task instruction only) and \emph{curated Skills} (the task's full \texttt{environment/skills/} directory). A third \emph{self-generated Skills} condition---the agent first authors skill packs with Anthropic's \texttt{skill-creator}, then solves with only those packs---is evaluated on the three dedicated-harness configurations and reported separately (Appendix~\ref{app:self-gen}).

\textbf{Evaluation harness and protocol.}
We instantiate \benchmarkName{} on \textbf{BenchFlow}~\citep{benchflowteam2026benchflow}, an open-source multi-turn agent benchmarking framework with uniform Skill injection, sandboxing, and trajectory logging. For each (configuration, task, condition) triple, BenchFlow builds a fresh pinned container, hands the agent the task instruction (and Skills under the relevant condition), and runs until final submission. The deterministic \texttt{test-script} verifier then emits a pass/fail result. Unscored, stale, or rate-limited trajectories are treated as incomplete coverage for rerun/audit; timeout rows are used only when healthy pass/fail replacements are unavailable and then scored as failures. The latest aggregate targets three selected public trials per (configuration, task, condition) cell against the fixed $87\times3$ frame.

\textbf{Metrics.}
\label{sec:metrics}
\label{sec:experimental-setup}
The primary metric is task-macro pass rate, following Terminal-Bench~\citep{merrill2026terminalbenchbenchmarkingagentshard}: per-task pass/fail outcomes are averaged over the three-trial frame, then across the fixed 87-task inventory. We complement this with normalized gain~\citep{hake1998interactive}
\begin{equation}
g \;=\; (\text{pass}_{\text{skill}} - \text{pass}_{\text{vanilla}}) \,/\, (1 - \text{pass}_{\text{vanilla}}),
\label{eq:normalized_gain}
\end{equation}
which measures the fraction of remaining headroom Skills close. Because every condition runs on the same task in the same container, deltas are paired differences at the (configuration, task) level, not unpaired-pool differences. Aggregation rules and per-cell trial counts are in Appendix~\ref{app:experimental-protocol-details}. For rows that summarize multiple configurations, we compute $g$ separately for each configuration at full precision and macro-average those normalized gains, so the reported mean $g$ can differ slightly from recomputing $g$ from mean pass rates.

            \section{Results}
\label{sec:results}

We report aggregate no-Skills vs.\ curated-Skills efficacy, then analyze Skills design factors, domain effects, and task-level failures.
\subsection{Skills Efficacy Across Model--Harness Combinations}

We evaluate each commercial model--harness configuration under matched no-Skills and curated-Skills conditions on all 87 tasks.

\subsubsection{Main Results}

\autoref{tab:main-results} reports pass rates for each model--harness combination, ordered by with-Skills performance.

\begin{table}[!t]
\centering
\small
\caption{Pass rates (\%), absolute gain ($\Delta$, pp), and normalized gain ($g$, \%) across the latest 87-task no-Skills vs.\ curated-Skills aggregate. $\Delta$ and $g$ are computed at full precision; configurations are ordered by Curated-Skills pass rate. The Mean-row $g$ macro-averages per-configuration normalized gains rather than recomputing $g$ from the mean pass rates. Each pass cell uses the fixed $87\times3$ trial frame; per-cell public coverage is reported in Appendix~\ref{app:results-summary}. Differences from rounded \emph{Pass} cells may differ by $\pm 0.1$\,pp.}
\label{tab:main-results}
\vspace{0.45em}
\begin{threeparttable}
\setlength{\tabcolsep}{4.5pt}
\renewcommand{\arraystretch}{1.02}
\begin{tabular}{llrrrr}
\toprule
& & \multicolumn{1}{c}{\textbf{No Sk.}}
& \multicolumn{3}{c}{\textbf{Curated Skills}} \\
\cmidrule(lr){3-3} \cmidrule(lr){4-6}
\textbf{Harness} & \textbf{Model}
& \textbf{Pass}
& \textbf{Pass} & \textbf{$\Delta$\,(pp)} & \textbf{$g$\,(\%)} \\
\midrule
OpenHands   & GPT-5.5                 & 51.5 & 67.3 & \ppos{15.8} & 32.6 \\
Codex       & GPT-5.5                 & 46.8 & 66.5 & \ppos{19.7} & 37.0 \\
Claude Code & Opus 4.7                & 43.0 & 61.2 & \ppos{18.2} & 31.9 \\
Gemini CLI  & Gemini 3.1 Pro          & 36.0 & 60.8 & \ppos{24.8} & 38.7 \\
OpenHands   & GLM 5.1                 & 32.7 & 58.4 & \ppos{25.7} & 38.1 \\
OpenHands   & Claude Opus 4.8         & 45.7 & 54.1 & \ppos{8.4} & 15.5 \\
OpenHands   & Kimi K2.6               & 33.4 & 54.0 & \ppos{20.6} & 31.0 \\
OpenHands   & Claude Opus 4.7         & 42.1 & 53.1 & \ppos{11.1} & 19.1 \\
OpenHands   & MiniMax M3              & 29.7 & 53.0 & \ppos{23.3} & 33.2 \\
OpenHands   & Gemini 3.1 Pro          & 33.8 & 52.8 & \ppos{19.0} & 28.7 \\
OpenHands   & DeepSeek V4 Pro         & 26.9 & 50.1 & \ppos{23.2} & 31.8 \\
OpenHands   & Gemini 3.5 Flash        & 41.1 & 48.2 & \ppos{7.1} & 12.1 \\
OpenHands   & Claude Sonnet 4.6       & 33.5 & 47.2 & \ppos{13.6} & 20.5 \\
OpenHands   & DeepSeek V4 Flash       & 27.5 & 44.7 & \ppos{17.2} & 23.7 \\
OpenHands   & Grok 4.3                & 22.8 & 41.7 & \ppos{18.8} & 24.4 \\
OpenHands   & GPT-5.4 Mini            & 29.9 & 41.4 & \ppos{11.5} & 16.4 \\
OpenHands   & MiniMax M2.7            & 18.1 & 34.9 & \ppos{16.8} & 20.5 \\
OpenHands   & Gemini 3.1 Flash Lite   & 16.0 & 20.1 & \ppos{4.1} & 4.9 \\
\midrule
\textbf{Mean} & & 33.9 & 50.5 & \textbf{\ppos{16.6}} & 25.5 \\
\bottomrule
\end{tabular}
\end{threeparttable}
\vskip -0.05in
\end{table}

\paragraph{Finding 1: Skills provide broad but non-uniform gains.} Every one of the 18 model--harness configurations improves with curated Skills (Figure~\ref{fig:skills_results}). The average gain is +16.6\,pp, but the spread is large (+4.1 to +25.7\,pp): most configurations see double-digit gains, while OpenHands + Gemini~3.1 Flash Lite is nearly flat (+4.1\,pp). Skill efficacy is therefore an empirical property of a specific agent stack rather than a universal constant.

\paragraph{Finding 2: The strongest absolute systems are not always the largest beneficiaries.} The highest with-Skills pass rates are OpenHands + GPT-5.5 (67.3\%), Codex + GPT-5.5 (66.5\%), and Claude Code + Opus~4.7 (61.2\%). The largest lifts instead come from OpenHands + GLM~5.1 (+25.7\,pp; 38.1\% normalized gain), Gemini CLI + Gemini~3.1 Pro (+24.8\,pp), and OpenHands + DeepSeek~V4 Pro (+23.2\,pp). Conversely, OpenHands + Gemini~3.5 Flash starts from a strong no-Skills baseline (41.1\%) but gains only +7.1\,pp, showing that high base capability does not imply high Skill leverage.

\paragraph{Finding 3: Harness choice materially changes how the same model uses Skills.} Fig.~\ref{fig:skills_results} includes three model families evaluated under multiple harnesses. GPT-5.5 reaches 67.3\% with Skills in OpenHands and 66.5\% in Codex; Gemini~3.1 Pro reaches 60.8\% in Gemini CLI but 52.8\% in OpenHands; Claude Opus~4.7 reaches 61.2\% in Claude Code and 53.1\% in OpenHands. Skills are therefore not merely extra context: the harness's discovery, prompting, execution, and tool-use loop mediate realized benefit. Appendix~\ref{app:results-summary} reports the per-configuration public-trial counts.

\begin{figure}[!t]
\centering
\includegraphics[width=\textwidth]{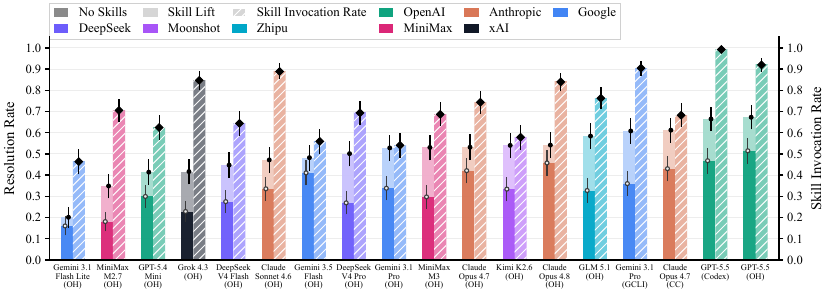}
\caption{Task-specific Skill Invocation Rate alongside resolution rates. Left axis: stacked resolution-rate bars (no-Skills baseline plus curated-Skills lift). Right axis: adjacent hatched Skill Invocation Rate bars. Open/filled circles mark 95\% CIs for no-Skills/with-Skills resolution rates; black diamonds mark invocation-rate CIs.}
\label{fig:skill-invocation-rate}
\end{figure}

\paragraph{Finding 4: Skill discovery is usually not the bottleneck.}
\autoref{fig:skill-invocation-rate} separates harness-bundled Skills, treated as part of the harness, from task-specific Skills shipped with each benchmark task. We count only the latter: a curated-Skills trial counts when its trajectory reads or invokes a Skill under that task's \texttt{environment/skills}. High Skill Invocation Rate does not guarantee high resolution, so the remaining failures often occur after task-Skill access. Exact counts are in Appendix~\ref{app:skill-invocation-rate}.

\paragraph{Self-generated Skills do not substitute for curated ones.} On the three dedicated-harness configurations we additionally evaluate a self-generated condition: the agent first authors skill packs with Anthropic's \texttt{skill-creator}, then solves each task with only those packs (Appendix~\ref{app:self-gen}). Self-generated Skills land \emph{below} the no-Skills baseline on all three configurations ($-8.1$\,pp on Claude Code + Opus~4.7, $-11.3$\,pp on Codex + GPT-5.5, $-11.5$\,pp on Gemini CLI + Gemini~3.1~Pro), while curated Skills add $+18.2$ to $+24.8$\,pp on the same configurations. A trajectory audit attributes the deficit to generated packs the solver never discovers, creator-side authoring that displaces solver work, and confidently wrong pack content when the skills are used (Appendix~\ref{app:self-gen-audit}).

\subsubsection{Domain-Level Analysis}

\begin{table}[!t]
\centering
\small
\caption{Skills efficacy (\%) by domain across the latest 87-task benchmark set. $N$ is the number of tasks per domain. Pass rates use a fixed three-trial denominator within each (condition, config, task), then macro-average across tasks and 18 model--harness configurations; all domains show positive aggregate delta.}
\label{tab:domain}
\vspace{0.45em}
\setlength{\tabcolsep}{5.5pt}
\renewcommand{\arraystretch}{1.03}
\begin{tabular}{lcccc}
\toprule
\textbf{Domain} & \textbf{N} & \textbf{No Skills} & \textbf{With Skills} & \textbf{$\Delta_\text{abs}$} \\
\midrule
Natural Science                & 14 & 42.0\% & 70.8\% & \ppos{28.8} \\
Media \& Content Production    &  5 & 23.3\% & 47.4\% & \ppos{24.1} \\
Cybersecurity                  &  7 & 29.5\% & 48.4\% & \ppos{18.9} \\
Industrial \& Physical Systems & 14 & 23.9\% & 39.6\% & \ppos{15.7} \\
Finance \& Economics           &  9 & 19.1\% & 33.3\% & \ppos{14.2} \\
Office \& White Collar         & 14 & 40.5\% & 53.0\% & \ppos{12.6} \\
Software Engineering           & 16 & 37.6\% & 49.2\% & \ppos{11.6} \\
Mathematics \& OR              &  8 & 45.7\% & 55.4\% & \ppos{9.7} \\
\bottomrule
\end{tabular}
\vskip -0.05in
\end{table}

\paragraph{Finding 5: Skills benefit varies widely across domains.}
\autoref{tab:domain} shows substantial heterogeneity. Natural Science (+28.8\,pp), Media \& Content Production (+24.1\,pp), and Cybersecurity (+18.9\,pp) benefit most; Software Engineering (+11.6\,pp) and Mathematics \& OR (+9.7\,pp) benefit least. Domains requiring specialized procedural knowledge underrepresented in model pretraining (e.g., scientific signal processing, security analysis, and multimedia transformation workflows) improve most, whereas domains with stronger pretraining and tooling coverage benefit less from external procedural guidance.

\subsubsection{Task-Level Analysis}

Task-level results reveal high variance in Skills effectiveness:

\paragraph{Top Skills beneficiaries.} The 10 highest-$\Delta$ tasks span prefix-cache replay, intrusion detection, SEC filings, wet-lab analysis, hydrology, geoscience, 3D parsing, optimization, and dependency auditing, with mean improvement $+67.0$\,pp. \nolinkurl{llm-prefix-cache-replay} reaches 94.4\% from a 1.9\% no-Skills pass rate, and \nolinkurl{dapt-intrusion-detection} reaches 81.5\% from zero (full list in Appendix~\ref{app:skills-delta}; Skill content audit in Appendix~\ref{app:skill-author-patterns}).

\paragraph{Failure modes when Skills hurt performance.} 13 of 87 tasks show negative Skills deltas; the largest drops occur on \nolinkurl{exam-block-sequencing} ($-7.4$\,pp), \nolinkurl{suricata-custom-exfil} ($-7.4$\,pp), and \nolinkurl{adaptive-cruise-control} / \nolinkurl{mars-clouds-clustering} / \nolinkurl{r2r-mpc-control} / \nolinkurl{econ-detrending-correlation} ($-5.6$\,pp each). Paired-trajectory audit (Appendix~\ref{app:skill-failure-modes}) surfaces three repeatable patterns: the Skill prescribes an unnecessarily heavyweight pipeline, displaces a stronger default strategy, or points the agent at a solver it cannot debug. The common cause is a single ``correct'' pipeline without applicability boundaries or lightweight fallbacks.

\paragraph{Finding 6: Compact, focused Skills outperform exhaustive ones, and small models with Skills can match larger models without.}
Two design ablations (Appendix~\ref{app:skill-design-factors}) refine the picture. First, Skill \emph{quantity}: tasks paired with one Skill gain +18.0\,pp, 2--3 Skills gain +19.0\,pp, and $\geq4$ Skills give only +10.1\,pp, suggesting excess content creates overhead or conflicting guidance. Second, Skill \emph{complexity}: compact and standard-length Skills (+19.0 and +21.5\,pp) outperform detailed (+14.5\,pp) and comprehensive documentation (+0.7\,pp); focused procedural guidance beats exhaustive prose. Finally, scale: MiniMax~M2.7 with Skills (34.9\%) exceeds stronger no-Skills baselines such as OpenHands + GLM~5.1 (32.7\%) and OpenHands + MiniMax~M3 (29.7\%), so Skills can partly compensate for model capacity on procedural tasks.

\vspace{-4pt}
\section{Discussion}
\vspace{-4pt}
\label{sec:discussion}

\textbf{Skills close procedural gaps.}
\looseness=-1
Skills are most helpful when success depends on concrete procedures and verifier-facing details (steps, constraints, sanity checks), rather than broad conceptual knowledge. Gains are largest for specialized workflows or brittle formats, and smaller or negative when models already have strong priors or the Skill adds overhead.

\textbf{Harnesses mediate Skills use.}
\looseness=-1
Skills efficacy depends not only on Skills quality but also on harness implementation. Some harnesses reliably retrieve and use Skills, while others acknowledge Skills content but proceed without invoking it. Structured interfaces can also introduce long-trajectory failure modes (e.g., format drift), reducing the influence of early-injected Skills. This motivates evaluating Skills under multiple harnesses rather than treating ``with Skills'' as a single condition.

\textbf{Implications for Skill authoring.}
Trajectory audit on the 10 highest-$\Delta$ tasks (Appendix~\ref{app:skill-author-patterns}) yields five recurring patterns: executable scripts with calibrated defaults, canonical data sources and parsing quirks, verifier-facing file-format constraints, algorithmic invariants, and task-specific \texttt{description:} frontmatter for first-scan matching. Comprehensive prose is essentially flat while focused documentation yields larger aggregate lift ($+0.7$\,pp vs.\ +19.0/+21.5\,pp; Appendix~\ref{app:skill-design-factors}); Skill authors should optimize for verifier-facing detail an agent cannot infer, not for completeness.

\textbf{Complexity-aware fallback paths for Skills.} The three failure modes that drive negative deltas (Appendix~\ref{app:skill-failure-modes}) all share the same root cause: the Skill prescribes a pipeline that is correct in principle but too heavy or brittle for routine agent execution. We propose extending the \texttt{SKILL.md} frontmatter with an explicit complexity contract: expected tool and token cost, applicability boundaries, and a required lightweight fallback path. This would help agents route around heavy recipes when the task evidence does not justify them, and it would force Skill authors to state when their ``correct'' workflow assumes unusually expensive computation.

\subsection{Limitations and Future Work}

\textbf{Coverage and generalization.}
\looseness=-1
\benchmarkName{} focuses on terminal-based, containerized tasks for reproducible evaluation, so results may not transfer directly to GUI agents, multi-agent coordination, or very long-horizon workflows. We also evaluate a limited set of models and harnesses whose Skills integration can change over time. A natural extension is multi-modal Skills and protocols for vision-language agents in GUI environments.


\textbf{Causal attribution and controls.}
Skills injection increases context length, so observed gains could partly reflect ``more context'' rather than procedural structure.
Our self-generated Skills condition suggests that model-authored procedural text does not reproduce the curated-Skills gain (Appendix~\ref{app:self-gen}), though its deficit mixes content quality with skill-discovery and creator/solver interference effects. Future work requires stronger length-matched baselines, such as random or irrelevant text and retrieval-only documentation, to isolate which components (steps, examples, code resources) drive improvement and to study automatic Skills synthesis.


\textbf{Determinism, contamination, and ecological validity.}
\looseness=-1
Containerization provides state isolation but not perfect determinism or immunity to training-set leakage.
We mitigate with multiple runs, a leakage audit (\S\ref{sec:SkillsBench}), and paired (Skills vs.\ no Skills) comparisons, yet cannot eliminate all nondeterminism or memorization effects.
Future work should evaluate ecosystem-representative settings, including lower-quality and automatically selected Skills, and study Skills composition: when multiple Skills help or interfere, and whether composite performance can be predicted from atomic effects.

\vspace{-4pt}
\section{Related Work}
\vspace{-4pt}
\label{sec:related-work}

We situate \benchmarkName{} within agent benchmarks, procedural augmentation, and evaluation methodology.

\textbf{Agent benchmarks.}
\looseness=-1
Recent benchmarks evaluate end-to-end agent capability across realistic environments, including Terminal-Bench~\citep{merrill2026terminalbenchbenchmarkingagentshard}, SWE-bench and follow-ons~\citep{jimenez2024swebench,yang2024sweagent,yang2025swesmith}, AgentBench and web/GUI settings~\citep{liu2023agentbench,zhou2024webarena,koh2024visualwebarena,xie2024osworld}, and suites for tool-mediated workflows, execution feedback, or domain specialization~\citep{yao2025taubench,trivedi2024appworld,yang2023intercode,chan2025mlebench,zhang2025cybench,zhuo2025bigcodebench,austin2021programsynthesis,ye2025replicationbenchaiagentsreplicate}.
These benchmarks measure fixed-agent task completion. \benchmarkName{} instead measures \emph{augmentation efficacy} via paired evaluation.


\textbf{Procedural augmentation and tool use.}
\looseness=-1
Prior work augments agents with structured reasoning or external knowledge, including CoALA and Voyager~\citep{sumers2024cognitive,wang2023voyager}, multi-step reasoning methods~\citep{wei2022chainofthought,yao2023tree,yao2023reactsynergizingreasoningacting,shinn2023reflexion,madaan2023selfrefine,zhou2023leasttomost,zhou2024lats}, retrieval and tool use~\citep{lewis2021retrievalaugmentedgenerationknowledgeintensivenlp,zhou2023docprompting,schick2023toolformerlanguagemodelsteach,qin2024toolllm}, and declarative optimization frameworks~\citep{khattab2023dspycompilingdeclarativelanguage}.
Skills combine procedural guidance with executable resources (\S\ref{sec:Back}); \benchmarkName{} quantifies their actual impact.


\textbf{Skills ecosystems and evaluation methodology.}
\looseness=-1
Anthropic's Agent Skills and MCP specifications~\citep{anthropic2025agentskills,anthropic2024mcp} formalized Skill packages and tool connectivity, and agent CLIs provide real-world harnesses~\citep{anthropic2025claudecode,google2025geminicli,openai2025codexcli}. \benchmarkName{} runs on BenchFlow~\citep{benchflowteam2026benchflow} and uses Terminal-Bench scoring~\citep{merrill2026terminalbenchbenchmarkingagentshard} for comparability.

\textbf{Paired evaluation of agent augmentations.}
\enlargethispage{1.5\baselineskip}
The closest methodological precedent is paired evaluation of inference-time augmentations on enterprise API workflows, which contrasts with-vs.-without conditions on the same task and reports paired bootstrap CIs~\citep{liu2023agentbench}. \benchmarkName{} adopts this paired-condition design while decoupling tasks from independently authored Skills and spanning 8 expertise-heavy domains rather than a single workflow class. Broader benchmarking practice motivates careful reporting and comparability~\citep{mattson2020mlperftrainingbenchmark,chiang2024chatbotarenaopenplatform,srivastava2023bigbench}; we therefore report both absolute gains and normalized gain~\citep{hake1998interactive} across No-Skills baselines (\S\ref{sec:metrics}).

\vspace{-4pt}
\section{Conclusion}
\vspace{-2pt}
\label{sec:conclusion}


We introduced \benchmarkName{}, a paired benchmark for evaluating Agent Skills as first-class artifacts. Across the latest 87-task, 18-configuration no-Skills vs.\ curated-Skills aggregate, curated Skills improve performance substantially but unevenly (+16.6\,pp on average; range +4.1\,pp to +25.7\,pp by configuration). Concise procedural Skills outperform exhaustive documentation, and Skills can partially substitute for model scale on procedural tasks. We release the benchmark, the BenchFlow harness, and the public trajectories/results.


\section*{Acknowledgments}
We thank the organizations whose models we evaluate---OpenAI, Anthropic, Google, Z.ai, Moonshot AI, DeepSeek, xAI, MiniMax, Alibaba Qwen, Tencent Hunyuan, and Xiaomi MiMo---for model access and API support. We thank Google Cloud Platform and Amazon Bedrock for cloud and inference infrastructure support, and the OpenHands team for the open-source harness used in our main aggregate. We thank Junxian He for guidance and advice; Yi R.\ (May) Fung for advice and paper-writing guidance; Gabriel Chua for providing last-minute API keys; Zachary Mueller for providing GPU resources through Lambda; and Ivan Burazin for providing Daytona credits. We also thank the following contributors for task and code contributions: Jianheng Hou, Jierun Chen, Jiahao Liu, Shutong Wu, Yulin Li, Zhenheng Tang, Benny Jiang, Qingyang Ma, Issa Sugiura, Jinya Jiang, Connor Adams, and Ananth Jayakrishnan Nambiar; and all open-source contributors to the \benchmarkName{} repositories.

\newpage
\bibliographystyle{plainnat}
\bibliography{references}

\appendix
\onecolumn

\section{Skill Ecosystem Analysis}
\label{app:ecosystem}

To contextualize \benchmarkName{} within the broader landscape of agent augmentation, we analyze the same construction snapshot reported in \autoref{fig:pipeline}: 234{,}896 direct GitHub/source Skills, 1{,}773{,}213 marketplace Skills, and 5{,}891 partner Skills, yielding 2{,}014{,}000 source-partitioned Skills (deduplicated within each source, then summed across partitions). When an analysis requires repository-level timestamps or cloned files, we report the reachable sample size explicitly while keeping 2{,}014{,}000 as the ecosystem-wide count.

\subsection{Data Collection}

We aggregate Skills from direct GitHub/source discovery, marketplace records, and partner inventories. Each entry records \texttt{name}, \texttt{description}, \texttt{author}, \texttt{url}, source partition, and, when available, the upstream repository \texttt{pushed\_at} timestamp. Summed across these source partitions (after deduplicating within each), the construction snapshot contains 2{,}014{,}000 source-partitioned Skills. This is a renewal of an earlier 37k-Skill snapshot taken in January 2026; the corpus has grown by more than an order of magnitude in the intervening months.

For per-Skill structural statistics (size, file count, extension mix), we further cloned 767{,}430 Skill bundles whose source repositories were still reachable; the remaining entries either point to deleted/private repositories, partner/marketplace records without cloneable source, or repositories that failed to fetch within retry limits.

\begin{figure}[htb]
  \begin{center}
    \centerline{\includegraphics[width=0.8\columnwidth]{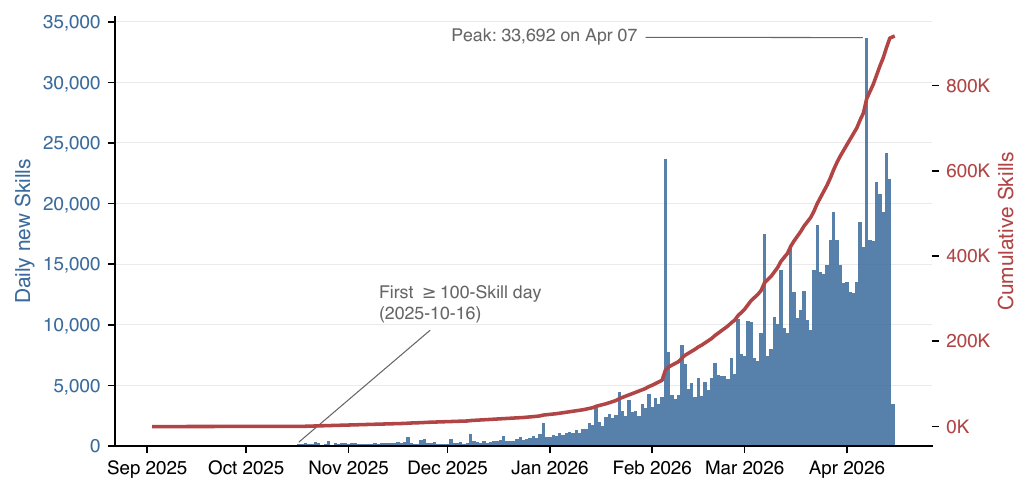}}
    \caption{Temporal dynamics of Skill creation in the timestamped portion of the 2{,}014{,}000-Skill construction snapshot. The window starts at 2025-09-03 to provide a pre-launch baseline. Daily additions (bars, left axis) climb through late 2025 and surge through Q1 2026, peaking at 33{,}692 Skills on 2026-04-07. The first day with $\geq$100 new Skills is 2025-10-16; the 153 Skills with timestamps before that date are likely artifacts of \texttt{pushed\_at} reflecting the upstream repo's last push (which can be backdated or simply inherited from a stale repo that later received a SKILL.md) rather than the Skill's actual creation time, so the pre-launch tail should not be read as evidence of real activity before the launch.}
    \label{fig:timeline}
  \end{center}
\end{figure}

\subsection{Skill Characteristics}

\begin{figure}[!t]
  \begin{center}
    \includegraphics[width=\linewidth]{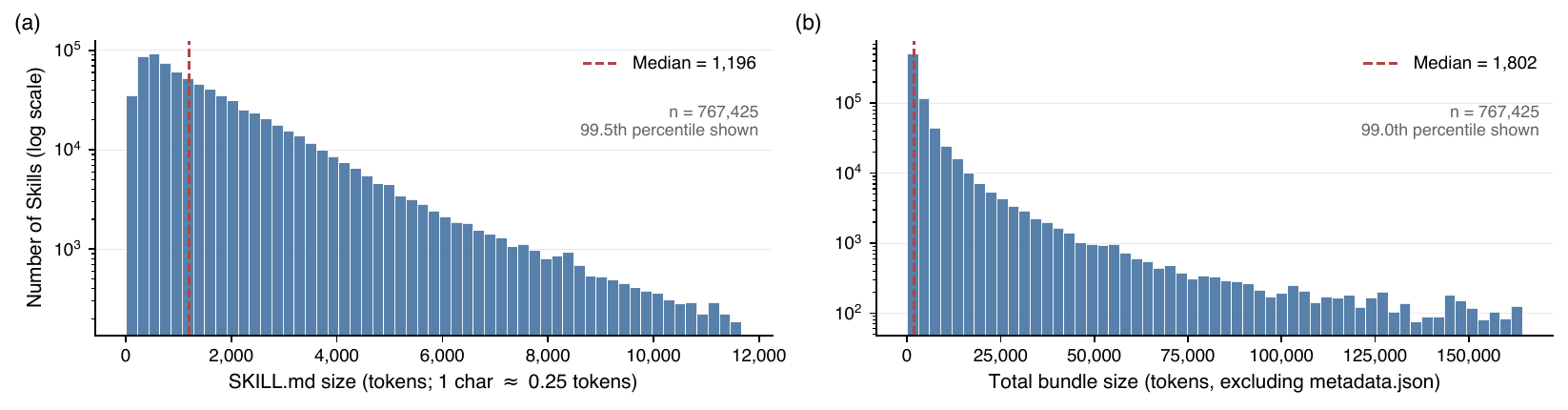}
    \caption{Skill size distributions in the reachable-clone sample (n=767{,}425; tokens approximated as bytes/4). \textbf{(a)} SKILL.md instructions, 99.5th percentile shown; median $\sim$1.2k tokens. \textbf{(b)} Total bundle size including all resources (excluding \texttt{metadata.json}), 99th percentile shown; median $\sim$1.8k tokens. Both distributions are highly skewed toward concise artifacts but the bundle-size tail extends several orders of magnitude.}
    \label{fig:sizes}
  \end{center}
\end{figure}

\paragraph{Size Distribution.}
Skill sizes follow a heavy-tailed log-normal distribution (\autoref{fig:sizes}). SKILL.md instructions (panel a) have a median of 4.8 KB ($\sim$1.2k tokens at 4 bytes/token; IQR 2.4--9.2 KB), while total bundle size (panel b) has a median of 7.2 KB ($\sim$1.8k tokens; IQR 3.0--17.4 KB). The tail is dramatic: the largest Skill bundle exceeds 1.6 GB.

\begin{figure}[htb]
  \begin{center}
    \centerline{\includegraphics[width=0.7\columnwidth]{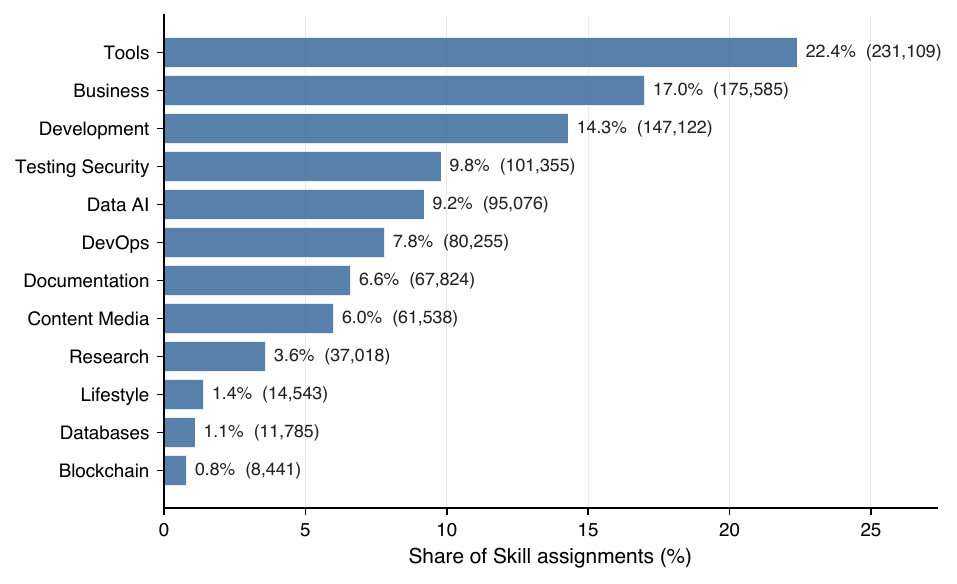}}
    \caption{Distribution of Skill categories across 1{,}031{,}651 marketplace assignments (snapshot 2026-04-30). No single category exceeds a quarter of the corpus, reflecting broad developer interest in software-engineering tooling, business workflows, testing/security, and data/AI.}
    \label{fig:category}
  \end{center}
\end{figure}

\paragraph{Domain Coverage.}
Skills span 12 marketplace-assigned categories with broad coverage and no single dominant area (\autoref{fig:category}; n=1{,}031{,}651 category assignments, where a Skill may carry multiple tags):
\begin{itemize}[nosep]
    \item Tools: 22.4\% (CLIs, terminal utilities, build tools)
    \item Business: 17.0\% (workflows, productivity, ops)
    \item Development: 14.3\% (general software engineering)
    \item Testing \& Security: 9.8\% (test scaffolds, security scans)
    \item Data \& AI: 9.2\% (data pipelines, ML, analytics)
    \item DevOps: 7.8\% (Docker, Kubernetes, CI/CD)
    \item Documentation: 6.6\% (technical writing, API docs)
    \item Content \& Media: 6.0\% (writing, design, media)
    \item Long tail: 6.9\% (Research, Lifestyle, Databases, Blockchain combined)
\end{itemize}

\begin{figure}[!t]
  \begin{center}
    \includegraphics[width=\linewidth]{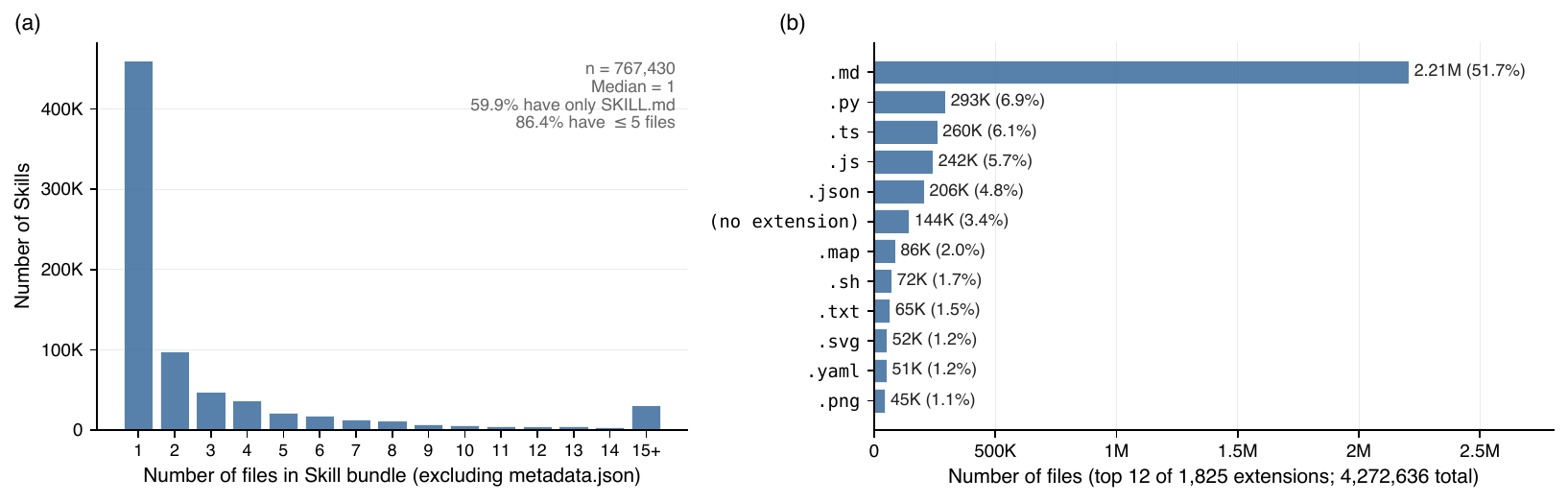}
    \caption{Structural patterns of Skill bundles in the reachable-clone sample. \textbf{(a)} File count per bundle (n=767{,}430, excluding \texttt{metadata.json}); most Skills are minimal, with 86.4\% containing five files or fewer. \textbf{(b)} Top 12 file extensions across the 4.27M files in the sample (out of 1{,}825 distinct extensions); markdown dominates at 51.7\%, with Python, TypeScript, JavaScript, and JSON as the next most common.}
    \label{fig:structure}
  \end{center}
\end{figure}

\paragraph{Structural Patterns.}
Most Skills with reachable cloned source are minimal (\autoref{fig:structure}, panel a): 59.9\% contain a single file (just \texttt{SKILL.md}) and 86.4\% contain five or fewer files (n=767{,}430, median 1, mean 5.6 inflated by the long tail). The file-extension mix (panel b) confirms the ecosystem is documentation-heavy: \texttt{.md} alone accounts for 51.7\% of all 4.27M files, followed by \texttt{.py} (6.9\%), \texttt{.ts} (6.1\%), \texttt{.js} (5.7\%), and \texttt{.json} (4.8\%). Natural-language instructions dominate over executable implementations.

\subsection{Quality Indicators}

We developed a quality scoring rubric based on:
\begin{enumerate}[nosep]
    \item \textbf{Completeness:} Presence of required components (0--3 points)
    \item \textbf{Clarity:} Readability and organization (0--3 points)
    \item \textbf{Specificity:} Actionable vs.\ vague guidance (0--3 points)
    \item \textbf{Examples:} Presence and quality of examples (0--3 points)
\end{enumerate}

Mean quality score across the ecosystem is 6.2/12 (SD=2.8), indicating substantial room for improvement in Skill authoring practices.

\subsection{Implications for Benchmark Design}

This ecosystem analysis directly informed \benchmarkName{} construction:
\begin{itemize}[nosep]
    \item \textbf{Domain selection:} Task categories mirror ecosystem coverage, ensuring Skills exist for evaluation
    \item \textbf{Quality awareness:} Ecosystem mean quality of 6.2/12 motivated our leakage audit and authoring guidelines---low-quality Skills would confound efficacy measurement
    \item \textbf{Skill selection:} We selected benchmark Skills from the top quality quartile (score $\geq$ 9/12) to isolate the effect of procedural knowledge from Skill quality variance
    \item \textbf{Size constraints:} Median Skill size ($\sim$1.8k tokens) informed our 8K context budget allocation
\end{itemize}

\paragraph{Limitation: Benchmark vs.\ Ecosystem Gap.}
Our 87 tasks with high-quality Skills represent an optimistic scenario.
Real-world Skill usage involves lower-quality Skills (ecosystem mean: 6.2/12 vs.\ benchmark mean: 10.1/12) and imperfect Skill-task matching.
Future work should evaluate with ecosystem-representative Skill samples.

                \section{Task Specification and Review Process}
\label{app:task-spec}

This appendix provides full details of the task specification format and quality control process summarized in \S\ref{sec:SkillsBench}. Every task is packaged according to the \texttt{task.md} standard---a single schema-validated document plus the role-named \texttt{verifier/} and \texttt{oracle/} directories---whose full specification is given in \appautoref{app:taskmd-standard}; this section documents the layout as instantiated by \benchmarkName{} and the review process that gates admission to the benchmark.

\subsection{Task Directory Structure}

Each task is a self-contained directory with the following layout:

\begin{lstlisting}[basicstyle=\footnotesize\ttfamily]
tasks/<task-id>/
  task.md                 # YAML frontmatter + task instruction
  environment/
    Dockerfile            # Container setup
    skills/               # Curated Skills (absent in no-Skills)
      <skill-name>/
        SKILL.md          # Required per skill
        scripts/          # Optional executable code
        references/       # Optional reference documentation
  oracle/
    solve.sh              # Oracle solution (must pass 100%)
  verifier/
    test.sh               # Runs pytest inside the container
    test_outputs.py       # Programmatic assertions
\end{lstlisting}

\subsection{Task Configuration (\texttt{task.md} Frontmatter)}

Each task specifies metadata and resource limits in the YAML frontmatter of \texttt{task.md}; the task instruction is the document body (\appautoref{app:taskmd-standard}):

\begin{lstlisting}[basicstyle=\footnotesize\ttfamily]
---
schema_version: "1.3"
metadata:
  author_name: Contributor Name
  author_email: email@example.com
  difficulty: medium      # easy | medium | hard
  category: finance-economics
  tags: [pandas, data-analysis, spreadsheet]
verifier:
  type: test-script
agent: {}
environment:
  network_mode: no-network
  cpus: 1                 # 1-4 cores
  memory_mb: 4096         # 2048-10240 MB
  storage_mb: 10240       # 10 GB standard
---
\end{lstlisting}

\subsection{Verification Infrastructure}

Verification typically uses pytest with the CTRF (Common Test Report Format) output (85 of 87 tasks; the remaining two drive a Python checker script directly); every task's \texttt{test.sh} writes its reward to \texttt{/logs/verifier/reward.txt}. The canonical \texttt{test.sh} installs dependencies, runs pytest, and writes a binary reward in the minimal pass/fail case:

\begin{lstlisting}[language=bash,basicstyle=\footnotesize\ttfamily]
#!/bin/bash
pip3 install --break-system-packages pytest pytest-json-ctrf
mkdir -p /logs/verifier
pytest --ctrf /logs/verifier/ctrf.json \
       /verifier/test_outputs.py -rA -v
if [ $? -eq 0 ]; then
  echo 1 > /logs/verifier/reward.txt
else
  echo 0 > /logs/verifier/reward.txt
fi
exit 0
\end{lstlisting}

\noindent A task passes if and only if all assertions in \texttt{test\_outputs.py} succeed (reward = 1); otherwise the verifier records a fail (reward = 0).

\subsection{PR Review Process}

Each submitted task undergoes a multi-stage review:
\begin{enumerate}[nosep]
\item \textbf{Automated CI}: Structural validation (\texttt{bench tasks check}), oracle execution (\texttt{bench eval create}\allowbreak\texttt{ --agent oracle}, must pass 100\%), and AI-detection screening (GPTZero) on the \texttt{task.md} instruction body.
\item \textbf{Maintainer review}: Evaluates data validity, task realism, oracle quality, Skill quality, and anti-cheating robustness. Reviewers run benchmark experiments with and without Skills across multiple agents.
\item \textbf{Benchmark report}: For each task, reviewers produce a structured report documenting oracle results, agent pass rates with and without Skills, failure analysis, and a final verdict (approve, major changes needed, or reject).
\end{enumerate}

\noindent Of 400 candidate submissions from 142 contributors, 87 tasks passed all review stages and are included in the current benchmark inventory (22\% acceptance rate). The empirical study reports near-complete paired public trajectories for the 87-task evaluation.

\subsection{Contributor Checklist}

All task submissions must satisfy the following requirements before review:
\begin{itemize}[nosep]
\item[$\square$] The \texttt{task.md} instruction is human-written (not model-generated); verified via GPTZero and human review
\item[$\square$] Skills are error-free, factually correct, and generalizable to similar tasks
\item[$\square$] Task is realistic and grounded in professional workflows
\item[$\square$] Task requires domain knowledge and is significantly easier with Skills
\item[$\square$] Outputs are deterministic and verifiable with programmatic assertions
\item[$\square$] Test count is below 10 unless justified; tests cover distinct criteria
\item[$\square$] Oracle solution passes 100\% (\texttt{bench eval create --agent oracle})
\item[$\square$] Task tested with agent both with and without Skills
\end{itemize}

\subsection{Maintainer Review Policy}

Maintainers evaluate each submission against seven criteria:
\begin{enumerate}[nosep]
\item \textbf{AI detection}: Verify the \texttt{task.md} instruction and configuration are manually written using GPTZero and human review. PRs with intentional grammar errors designed to circumvent AI detectors are closed.
\item \textbf{Data quality}: Data must be real-world and appropriately complex. AI-generated or toy data is rejected.
\item \textbf{Task validity}: Tasks must be grounded in realistic professional scenarios. Artificially inflated complexity is rejected.
\item \textbf{Oracle quality}: Simple solutions (e.g., an Excel formula or short script) are preferred over over-engineered oracle implementations.
\item \textbf{Author history}: Authors flagged multiple times across PRs are closed automatically.
\item \textbf{Test parsimony}: Fewer than 10 test cases unless justified; tests should cover distinct criteria rather than repeat similar checks.
\item \textbf{Multimodal verification}: For multimodal tasks (audio, PPTX, video, PDF), maintainers personally inspect agent output to verify correctness beyond programmatic assertions.
\end{enumerate}

\subsection{Automated CI Pipeline}

The CI pipeline performs the following checks on each PR:
\begin{itemize}[nosep]
\item \textbf{Structural validation} (\texttt{bench tasks check}): Verifies required files exist, the \texttt{task.md} frontmatter schema is valid, Dockerfile builds, and test structure is correct.
\item \textbf{Oracle execution} (\texttt{bench eval create --agent oracle}): Runs the oracle solution end-to-end and requires 100\% test pass rate.
\item \textbf{AI-detection screening}: Runs GPTZero on the \texttt{task.md} instruction body to flag potential model-generated content.
\item \textbf{LLM-backed quality checks}: Automated verification of behavior consistency between instructions and tests, anti-cheating measures, pinned dependency checks, typo detection, and hardcoded solution detection.
\end{itemize}

\subsection{Benchmark Report Template}

For each task, reviewers produce a structured report documenting:
\begin{enumerate}[nosep]
\item \textbf{Task metadata}: Name, category, difficulty, tags, description, Skills provided, key requirements.
\item \textbf{Oracle results}: Pass/fail status, reward, tests passed, timing.
\item \textbf{Agent results}: Pass rates per agent-model combination, with and without Skills, including execution time.
\item \textbf{Skills impact}: Quantified comparison of with-Skills vs.\ without-Skills performance per agent.
\item \textbf{Failure analysis}: Per-test breakdown of failures including actual vs.\ expected output, root cause, and evidence from trajectories.
\item \textbf{Recommendation}: One of: \textsc{approve}, \textsc{approve with caveats}, \textsc{major changes needed}, or \textsc{reject}.
\end{enumerate}

\subsection{Review Lifecycle}

PRs progress through a defined label-based lifecycle:
\begin{enumerate}[nosep]
\item \textbf{WIP} $\rightarrow$ \textbf{Need review}: Author signals readiness for initial review.
\item \textbf{Need review} $\rightarrow$ \textbf{Reviewing}: A maintainer begins active review and benchmark experiments.
\item \textbf{Reviewing} $\rightarrow$ \textbf{Change requested / Major change needed / Critical change needed}: Issues identified; author must address.
\item \textbf{Change requested} $\rightarrow$ \textbf{Take another look}: Author responds after changes.
\item \textbf{Ready to merge} $\rightarrow$ \textbf{Good task}: All reviews passed; task included in benchmark.
\end{enumerate}

\noindent Critical changes include unrealistic task scenarios, AI-generated instructions, or synthetic data. Major changes include incorrect tests, unreliable verifiers, or poor Skill quality requiring re-evaluation.

\subsection{Task Quality Criteria}

Tasks are evaluated against the following criteria:
\begin{itemize}[nosep]
\item \textbf{Realistic}: Grounded in professional workflows that people in that domain actually perform
\item \textbf{Skill-dependent}: Significantly easier with Skills than without---tasks solvable without any procedural guidance are rejected
\item \textbf{Verifiable}: Deterministic outputs testable with programmatic assertions; LLM-as-judge is not used
\item \textbf{Composable}: Tasks should exercise 3--6 Skills together; instructions never reference which Skills to use
\item \textbf{Test parsimony}: Fewer than 10 test cases unless justified; tests should cover distinct criteria rather than repeat similar checks
\end{itemize}

\section{The \texttt{task.md} Task Standard}
\label{app:taskmd-standard}

We claim in \S\ref{sec:intro} that the paired-evaluation protocol is reusable beyond our own task set: practitioners can run the same harness on their own Skill libraries before shipping. That reuse requires the task package to be portable and versioned. The \texttt{task.md} standard therefore packages a task as a single schema-validated document---YAML frontmatter for configuration, Markdown body for the task instruction---together with runtime directories named for their roles (\texttt{verifier/}, \texttt{oracle/}), so that all 87 evaluated tasks load through one parser. \appautoref{app:task-spec} shows the package layout as instantiated by \benchmarkName{}; this appendix specifies the standard itself.

Every pass rate reported in this paper comes from the deterministic \texttt{test-script} verifier described below. The standard is strictly more expressive than \benchmarkName{} requires: it defines verifier strategies the benchmark deliberately does not use (\autoref{tab:verifier-strategies}), so the format's generality does not come at the expense of the no-LLM-as-a-judge determinism on which the main measurement depends (\S\ref{sec:SkillsBench}). The enforcement properties stated below hold in the open-sourced \benchmarkName{} reference harness~\citep{benchflowteam2026benchflow} and are covered by its test suite. Beyond the reference harness, the standard is supported by the AgentBeats platform: tasks packaged as \texttt{task.md} run directly on AgentBeats, where \benchmarkName{} is publicly available.\footnote{\url{https://agentbeats.dev/Yiminnn/skillsbench-agentbeats}}

Although initially based on the Harbor task format \citep{harborframeworkteam2026harborframework}, \benchmarkName{} tasks required a new approach to support the self-generated Skills condition (\appautoref{app:self-gen}). This condition relies on two isolated agent sessions (a creator and a clean solver), whereas Harbor only supports single, continuous rollouts. To resolve this, we developed the \texttt{task.md} standard. This format extends Harbor's containerized framework by merging configuration and instruction files into a single schema-validated document, adding controlled metadata and explicit verifier strategies (\autoref{tab:verifier-strategies}), and renaming the \texttt{solution/} and \texttt{tests/} directories to \texttt{oracle/} and \texttt{verifier/} to better reflect their roles.

\subsection{Document Structure}

\texttt{task.md} is a Markdown file whose YAML frontmatter is the task configuration and whose body, immediately following the closing frontmatter delimiter, is the task instruction---no section heading is required. The schema is strict at the top level: unknown keys are rejected at parse time, so a contributor's task cannot silently lose a configuration field as the schema evolves across harnesses; values \emph{within} the free-form \texttt{metadata} block are open by design (next subsection).

\begin{lstlisting}[basicstyle=\footnotesize\ttfamily]
---
schema_version: "1.3"
metadata:
  difficulty: medium           # easy | medium | hard
  category: finance-economics  # controlled vocabulary
  task_type: [analysis]
  tags: [pandas, spreadsheet]  # open tier
environment:
  network_mode: no-network     # no-network|public|allowlist
  cpus: 1
  memory_mb: 4096
  storage_mb: 10240
agent: {}
verifier: {type: test-script}
oracle: {}
---
You are given ... produce ... in /app/out.json.
\end{lstlisting}

\subsection{Controlled Metadata Vocabulary}

The \texttt{metadata} block is two-tier. A \emph{closed} tier---\texttt{category}, \texttt{task\_type}, \texttt{modality}, \texttt{interface}, \texttt{skill\_type}, and \texttt{difficulty}---draws from controlled vocabularies, of which \texttt{category} and \texttt{difficulty} correspond to the benchmark taxonomy of \appautoref{app:domain-mapping}; an \emph{open} tier (\texttt{tags} and free-form Skill names) preserves authoring flexibility. The closed tier is enforced at review time rather than at parse time: \texttt{metadata} is parsed as a free-form mapping with no load-time enumeration check, so off-taxonomy values are surfaced during review rather than rejected by the parser. This keeps authoring lightweight while still yielding the benchmark-wide classification over which the per-domain breakdowns are computed.

\subsection{Explicit Network Policy}

Because uncontrolled network egress is a confound for a deterministic verifier, the standard makes egress explicit: access is declared with a single enumerated field, \texttt{network\_mode} $\in$ \{\texttt{no-network}, \texttt{public}, \texttt{allowlist}\}, rather than a coarse on/off boolean. The \texttt{allowlist} mode requires a non-empty \texttt{allowed\_hosts} list---a parse-time error otherwise---so an \texttt{allowlist} declaration is never left under-specified. A task whose data is baked into the container image declares \texttt{no-network}; a task that must reach a fixed external service declares \texttt{allowlist} with the specific hosts.

\subsection{Verifier Strategies}

A verifier strategy is selected by a \texttt{type} field, either in the \texttt{verifier} block of \texttt{task.md} or in an optional \texttt{verifier/verifier.md} manifest, where the same strategy is spelled \texttt{script} rather than \texttt{test-script} (\autoref{tab:verifier-strategies}). Absent a manifest, the configuration's \texttt{verifier.type} selects the default \texttt{test-script} path, which runs \texttt{verifier/test.sh} and reads a pass/fail result from \texttt{reward.txt} (or a structured \texttt{reward.json}); \texttt{test-script} is the only strategy the 87 tasks use, which keeps the measurement free of LLM-as-a-judge variance (\S\ref{sec:SkillsBench}). The remaining strategies---\texttt{llm-judge}, \texttt{reward-kit}, and \texttt{agent-judge}\,/\,\texttt{ors-episode}---exist for task classes whose correctness a deterministic script cannot decide; they are part of the standard but not of this benchmark. Unscored, stale, or rate-limited rows are treated as incomplete coverage; timeout rows enter only when healthy replacements are unavailable and are scored as failures.

\begin{table}[htb]
\centering
\small
\caption{Verifier strategies defined by the \texttt{task.md} standard. \benchmarkName{} uses \texttt{test-script} exclusively to preserve deterministic, programmatic verification.}
\label{tab:verifier-strategies}
\begin{tabular}{lll}
\toprule
\textbf{Strategy} & \textbf{Reward source} & \textbf{In \benchmarkName{}} \\
\midrule
\texttt{test-script} & \texttt{test.sh} $\rightarrow$ \texttt{reward.txt/json} & all 87 tasks \\
\texttt{llm-judge} & rubric-scored deliverables & not used \\
\texttt{reward-kit} & external reward entrypoint & not used \\
\texttt{agent-judge} & judged transcript & not used \\
\texttt{ors-episode} & episode evidence $\rightarrow$ reward & not used \\
\bottomrule
\end{tabular}
\end{table}

\section{Experimental Setup Details}
\label{app:exp-details}

\noindent This appendix expands the experimental setup summarized in \S\ref{sec:experimental-setup}.

\subsection{Setup overview}

We instantiate the \benchmarkName{} protocol across commercial terminal agents and model--harness configurations. The latest aggregate targets a fixed 9{,}396-slot frame (18 configurations $\times$ 87 tasks $\times$ 2 conditions $\times$ 3 trials), with 9{,}396 selected public result files currently available. All runs use temperature 0; unscored, stale, or rate-limited rows are tracked as coverage gaps for audit and rerun, while timeout rows are used only when healthy pass/fail replacements are unavailable and are scored as failures.

\paragraph{Models.}
The latest aggregate pairs OpenHands with GPT-5.5, GPT-5.4 Mini, Claude Opus~4.8/4.7, Claude Sonnet~4.6, Gemini~3.1 Pro, Gemini~3.5 Flash, Gemini~3.1 Flash Lite, GLM~5.1, Kimi~K2.6, DeepSeek~V4 Pro/Flash, Grok~4.3, and MiniMax~M3/M2.7; Gemini CLI with Gemini~3.1 Pro; Claude Code with Claude Opus~4.7; and Codex CLI with GPT-5.5. Full identifiers and selected result counts are listed in \autoref{tab:models-full}; protocol-level details appear in \appautoref{app:experimental-protocol-details}.

\paragraph{Metrics.}
Our primary metric is task-macro pass rate following Terminal-Bench~\citep{merrill2026terminalbenchbenchmarkingagentshard}: verifier-scored pass/fail outcomes are averaged over repeated trials within each task and then averaged across the fixed inventory of 87 tasks. We report absolute Skills improvement and normalized gain $g = (\text{pass}_{\text{skill}} - \text{pass}_{\text{vanilla}}) / (1 - \text{pass}_{\text{vanilla}})$ as defined in the main text (\S\ref{sec:metrics}).
We interpret $g$ alongside absolute pass-rate deltas to avoid ceiling-effect artifacts. Harness versions, resource limits, retry rules, and confidence-interval calculations are provided in \appautoref{app:experimental-protocol-details}.

\subsection{Model and Harness Configurations}

\autoref{tab:models-full} presents the 18 model--harness configurations evaluated in the latest aggregate.

\begin{table}[htb]
\centering
\footnotesize
\caption{Agent harnesses and models evaluated in the latest aggregate. Counts report selected no-Skills / curated-Skills public result files out of 261 possible per condition.}
\setlength{\tabcolsep}{3pt}
\begin{tabular}{llll}
\toprule
\textbf{Harness} & \textbf{Model} & \textbf{Provider} & \textbf{$n$} \\
\midrule
OpenHands & GPT-5.5 & OpenAI / Azure OpenAI & 261/261 \\
Codex & GPT-5.5 & OpenAI / Azure OpenAI & 261/261 \\
Claude Code & Opus 4.7 & Anthropic & 261/261 \\
Gemini CLI & Gemini 3.1 Pro & Google & 261/261 \\
OpenHands & GLM 5.1 & Z.ai / GLM & 261/261 \\
OpenHands & Claude Opus 4.8 & Anthropic / AWS Bedrock & 261/261 \\
OpenHands & Kimi K2.6 & Moonshot AI & 261/261 \\
OpenHands & Claude Opus 4.7 & Anthropic / AWS Bedrock & 261/261 \\
OpenHands & MiniMax M3 & MiniMax & 261/261 \\
OpenHands & Gemini 3.1 Pro & Google & 261/261 \\
OpenHands & DeepSeek V4 Pro & DeepSeek & 261/261 \\
OpenHands & Gemini 3.5 Flash & Google & 261/261 \\
OpenHands & Claude Sonnet 4.6 & Anthropic / AWS Bedrock & 261/261 \\
OpenHands & DeepSeek V4 Flash & DeepSeek & 261/261 \\
OpenHands & Grok 4.3 & xAI & 261/261 \\
OpenHands & GPT-5.4 Mini & OpenAI & 261/261 \\
OpenHands & MiniMax M2.7 & MiniMax & 261/261 \\
OpenHands & Gemini 3.1 Flash Lite & Google & 261/261 \\
\bottomrule
\end{tabular}

\label{tab:models-full}
\end{table}

\subsection{Harness Descriptions}

We evaluate four commercial agent harnesses:
\begin{itemize}[nosep]
    \item \textbf{OpenHands}: open-source terminal agent harness
    \item \textbf{Claude Code}~\citep{anthropic2025claudecode}: Anthropic's agent with native Skill integration
    \item \textbf{Gemini CLI}~\citep{google2025geminicli}: Google's open-source terminal agent
    \item \textbf{Codex CLI}~\citep{openai2025codexcli}: OpenAI's lightweight coding agent
\end{itemize}
These tightly couple specific models with proprietary agent logic, representing real-world deployment conditions.

\paragraph{Model Family Consideration.} Claude models have been trained with awareness of the Agent Skills specification~\citep{anthropic2025agentskills}, which may confer advantages when processing Skill-formatted instructions.

\subsection{Agent Interface}

Agents interact with the environment through a standardized interface:

\begin{lstlisting}[language=Python,basicstyle=\footnotesize\ttfamily]
class BaseAgent(ABC):
    @abstractmethod
    def step(self, obs: str) -> str:
        """obs: terminal output -> action"""
        pass
\end{lstlisting}

\vspace{-8pt}

\subsection{Skill Structure}

Each Skill is a directory containing a required \texttt{SKILL.md} file with YAML frontmatter and optional bundled resources:

\begin{lstlisting}[basicstyle=\footnotesize\ttfamily]
skill-name/
  SKILL.md          # Required: YAML frontmatter + instructions
  scripts/          # Optional: executable code
  references/       # Optional: reference documentation
\end{lstlisting}

\noindent The \texttt{SKILL.md} frontmatter specifies the Skill's name and a one-line description used by agents for skill discovery. The body contains procedural guidance, code examples, and usage patterns.

\subsection{Self-Generated Skills Condition}
\label{app:self-gen}

The self-generated condition asks whether an agent can replace curated Skills with Skills it authors for itself. Unlike the no-Skills and curated-Skills conditions, it consists of \emph{two isolated sessions} per task. In the \emph{creator} session, a clean agent receives the task package with all curated Skills removed and exactly one Skill mounted---Anthropic's official \texttt{skill-creator}---and is instructed to author skill packs rather than solve the task. The creator prompt, verbatim from the harness:\footnote{\texttt{/instruction.md} is the in-sandbox mount of the task instruction (the \texttt{task.md} body); \texttt{<task>} denotes the task name.}

\begin{quote}
\small\ttfamily
Use the skill-creator skill exactly as provided.

Read /instruction.md and inspect the task environment only as needed to understand the reusable workflow. Do not solve the task directly.

Create one or more complete Anthropic-standard skill packs as immediate child directories under: /app/generated-skills

Use this suggested path if one skill is enough: /app/generated-skills/<task>-skill

Each generated skill pack path must look like /app/generated-skills/<skill-name>/SKILL.md. It may include scripts/, references/, assets/, examples, or other bundled resources when they help a fresh solver avoid repeated work.

The solver context will start with a clean agent session and only the generated skill packs mounted. Make the skills useful for solving this task type from the same sandbox environment.
\end{quote}

\noindent In the \emph{solver} session, a fresh agent then runs the task exactly as in the curated-Skills condition, except that the curated Skills are replaced by the generated packs; the task instruction itself is unmodified, and none of the creator's reasoning is in the solver's context. The campaign realized this protocol in two ways: the Claude Code runs execute the creator and solver scenes back-to-back inside each trial's sandbox, regenerating the packs every trial, whereas the Codex and Gemini CLI runs generated the packs once per (task, configuration) and reused them across the scored solver trials. Because the Claude Code scenes share the task sandbox, file-system state can carry over beyond the pack files; the consequences of both realizations are examined in \appautoref{app:self-gen-audit}.

\paragraph{Configurations and accounting.} The condition is evaluated on the three dedicated-harness configurations of the main aggregate---Claude Code with Opus~4.7, Codex with GPT-5.5, and Gemini CLI with Gemini~3.1~Pro---using the same harness, model identifier, and sandbox as their baselines.\footnote{One deviation: the Claude Code self-generated runs used effort level \texttt{max}, whereas its baseline runs used \texttt{high}; the Codex (\texttt{xhigh}) and Gemini CLI configurations match their baselines exactly.} The self-generated condition is retained as a diagnostic rather than part of the main two-condition aggregate. The current audit reports verifier-scored trials against the same 87-task, three-trial frame: 258, 258, and 253 of the 261 slots are present for the three configurations respectively (the campaign skipped \texttt{pddl-airport-planning}); unscored or missing slots are treated as coverage gaps rather than accepted outcomes. The no-Skills and curated-Skills columns restate the main aggregate of \autoref{tab:main-results}.

\begin{table}[htb]
\centering
\small
\caption{Self-generated Skills vs.\ the matched baselines on the three dedicated-harness configurations (pass rate \%, fixed 87-task denominator). $\Delta_\text{G}$ = self-generated minus no Skills; $\Delta_\text{S}$ = curated Skills minus no Skills.}
\label{tab:selfgen}
\begin{tabular}{llrrrrr}
\toprule
\textbf{Harness} & \textbf{Model} & \textbf{No Sk.} & \textbf{Self-Gen} & \textbf{$\Delta_\text{G}$} & \textbf{Curated} & \textbf{$\Delta_\text{S}$} \\
\midrule
Claude Code & Opus 4.7       & 43.0 & 34.9 & \nneg{8.1}  & 61.2 & \ppos{18.2} \\
Codex       & GPT-5.5        & 46.8 & 35.5 & \nneg{11.3} & 66.5 & \ppos{19.7} \\
Gemini CLI  & Gemini 3.1 Pro & 36.0 & 24.5 & \nneg{11.5} & 60.8 & \ppos{24.8} \\
\bottomrule
\end{tabular}
\end{table}

\begin{figure}[htb]
\centering
\includegraphics[width=0.72\textwidth]{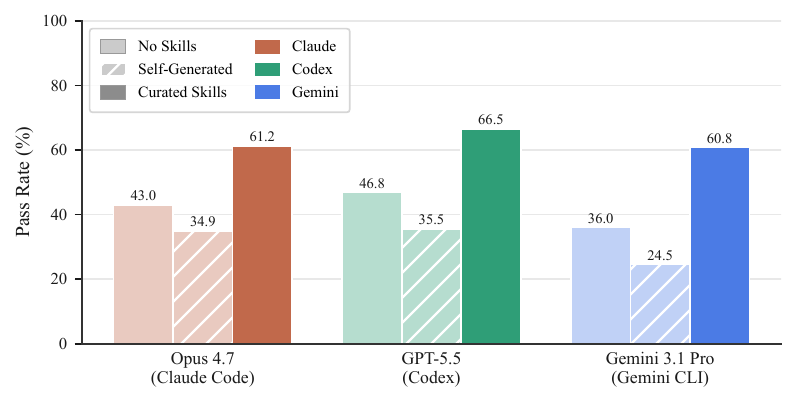}
\caption{Self-generated Skills underperform not only curated Skills but the no-Skills baseline on all three dedicated-harness configurations.}
\label{fig:selfgen}
\end{figure}

\noindent On all three configurations, self-generated Skills land \emph{below} the no-Skills baseline ($-8.1$ to $-11.5$\,pp) while curated Skills add $+18.2$ to $+24.8$\,pp (\autoref{tab:selfgen}, \autoref{fig:selfgen}). The trajectory audit below shows the deficit is not a single mechanism: generated packs frequently go unused by the solver, creator-side authoring can displace solver progress, and packs that are used can lock in confidently wrong assumptions.

\subsubsection{Why self-generated Skills underperform: trajectory audit}
\label{app:self-gen-audit}

We audited twelve solver trajectories across ten (task, configuration) pairs selected for extreme condition gaps---tasks where curated Skills pass but self-generated fail, the rare cases where self-generated beats no-Skills, and skill-dependent tasks---together with a protocol check comparing trials of the same task. Four mechanisms account for the audited gap:

\begin{itemize}[nosep,leftmargin=*]
\item \textbf{Generated packs frequently go unused.} In the pre-baked Codex and Gemini CLI family, 10 of 12 audited trajectories never list, read, or mention the generated skills---the event stream contains zero occurrences of ``skill''---so those passes \emph{and} failures are causally unrelated to pack content. Unlike curated Skills, which the harnesses surface through their native discovery mechanisms, a pack the solver never discovers can only be dead weight.
\item \textbf{Authoring displaces task work.} In the Claude Code realization the creator and solver scenes share a single trial context. On \texttt{threejs-to-obj}, creator-side work consumed the trajectory before a solver artifact was produced---trial logs end with the \texttt{SKILL.md} write still pending---even though the creator had already produced and verified a correct OBJ export (written only to \texttt{/tmp}, obeying ``Do not solve the task directly''). Both baselines pass this task; the self-generated miss is creator/solver interference, not knowledge failure.
\item \textbf{Consumed packs lock in confident errors.} When the solver does use a generated pack, wrong assumptions become load-bearing. On \texttt{3d-scan-calc}, the generated \texttt{SKILL.md} asserts as a ``critical gotcha'' that the STL coordinates are millimetres and hard-codes a $1000\times$ unit conversion---both solver trials ran the bundled script unquestioned and submitted a mass off by three orders of magnitude, with every other step correct. The curated Skill carries exactly the opposite warning (``Do not assume millimeters \ldots no unit conversion is needed''). On \texttt{radar-vital-signs} the generated pack replaced the curated multi-check pipeline with a fire-and-forget script whose plausibility bounds were too loose to flag its own implausible output, omitting the curated autocorrelation cross-check.
\item \textbf{The clearest win is leakage, not reuse.} The strongest positive case, \texttt{fix-visual-stability} (self-generated 1.0 vs.\ curated 0.0 on Claude Code), works because the creator wrote the pack \emph{inside the graded sandbox}: the generated \texttt{SKILL.md} names the app's exact offender components, enumerates every \texttt{data-testid} the verifier checks, and prescribes a fix order---an answer key for this instance rather than reusable procedure. The curated Skill ships only generic guidance plus measurement tooling.
\end{itemize}

\noindent Two caveats temper per-task readings (the aggregate direction in \autoref{tab:selfgen} is unaffected). First, some zeros are measurement artifacts: the \texttt{3d-scan-calc} verifier was patched ten days after these runs to also accept the millimetre interpretation, and unscored audit records are excluded from causal interpretation. Second, the audit's visibility is bounded: creator sessions for the pre-baked family are not part of the released artifacts, and Gemini CLI trajectories omit tool outputs, so non-discovery there is established behaviorally rather than from pack content.

\vspace{-8pt}

\subsection{Software and Model Versions}

\vspace{-8pt}

\autoref{tab:versions} lists representative model identifiers and harnesses used in the latest aggregate.

\vspace{-8pt}

\begin{table}[htb]
\centering
\footnotesize
\caption{Representative model API identifiers and harnesses in the latest aggregate.}
\label{tab:versions}
\begin{tabular}{lll}
\toprule
\textbf{Display Name} & \textbf{API Model ID} & \textbf{Harness Version} \\
\midrule
GPT-5.5 & \texttt{openai/gpt-5.5} & OpenHands / Codex CLI \\
GPT-5.4 Mini & \texttt{openai/gpt-5.4-mini} & OpenHands \\
Claude Opus 4.8 & \texttt{claude-opus-4-8} & OpenHands \\
Claude Opus 4.7 & \texttt{claude-opus-4-7} & OpenHands / Claude Code \\
Claude Sonnet 4.6 & \texttt{claude-sonnet-4-6} & OpenHands \\
Gemini 3.1 Pro & \texttt{gemini-3.1-pro-preview} & OpenHands / Gemini CLI \\
Gemini 3.5 Flash & \texttt{gemini-3.5-flash} & OpenHands \\
Gemini 3.1 Flash Lite & \texttt{gemini-3.1-flash-lite-preview} & OpenHands \\
GLM 5.1 & \texttt{glm/glm-5.1} & OpenHands \\
Kimi K2.6 & \texttt{kimi/kimi-k2.6} & OpenHands \\
Grok 4.3 & \texttt{xai/grok-4.3} & OpenHands \\
DeepSeek V4 Pro & \texttt{deepseek/deepseek-v4-pro} & OpenHands \\
DeepSeek V4 Flash & \texttt{deepseek/deepseek-v4-flash} & OpenHands \\
MiniMax M3 & \texttt{minimax/MiniMax-M3} & OpenHands \\
MiniMax M2.7 & \texttt{minimax/MiniMax-M2.7} & OpenHands \\
\bottomrule
\end{tabular}
\end{table}

\vspace{-8pt}

\subsection{Container Environment}

All tasks run in Docker containers built from an \texttt{ubuntu:24.04} base image. Per-task resource allocation is specified in the \texttt{environment} block of the task configuration (\appautoref{app:taskmd-standard}):

\begin{itemize}[nosep]
    \item \textbf{CPUs}: 1--4 cores (task-dependent)
    \item \textbf{Memory}: 2--10 GB (task-dependent)
    \item \textbf{Storage}: 10 GB (standard across all tasks)
    \item \textbf{GPU}: None (no tasks require GPU)
\end{itemize}

\noindent Containers are deleted after each trial (\texttt{delete: true}) to ensure no state leaks between runs.

\vspace{-8pt}

\subsection{Inference Configuration}

\begin{itemize}[nosep]
    \item \textbf{Temperature}: 0 (deterministic sampling)
    \item \textbf{Reasoning effort}: Highest available setting for each model
    \item \textbf{Context management}: Sliding window with 8K token limit; oldest turns dropped when exceeded
    \item \textbf{Run selection}: Public result files with complete metadata and healthy verifier-scored pass/fail outcomes are preferred; timeout rows are used only as failure backfill when healthy replacements are unavailable
\end{itemize}

\vspace{-8pt}

\subsection{Experiment Orchestration}

\begin{itemize}[nosep]
    \item \textbf{Runs per task}: three selected public result files for the main no-Skills and curated-Skills conditions
    \item \textbf{Concurrent trials}: 256 for the main conditions
    \item \textbf{Retry policy}: Unscored, stale, or rate-limited runs are audited and rerun
    \item \textbf{Task scope}: 87 tasks in the current released inventory
\end{itemize}

\noindent The latest aggregate tracks 9{,}396 fixed result slots across 18 configurations, 87 tasks, and 2 conditions; 9{,}396 public result files are currently selected.

\section{Task-Level Results}
\label{app:task-level}

\begin{figure}[p]
  \begin{center}
    \includegraphics[width=\textwidth,height=0.88\textheight,keepaspectratio]{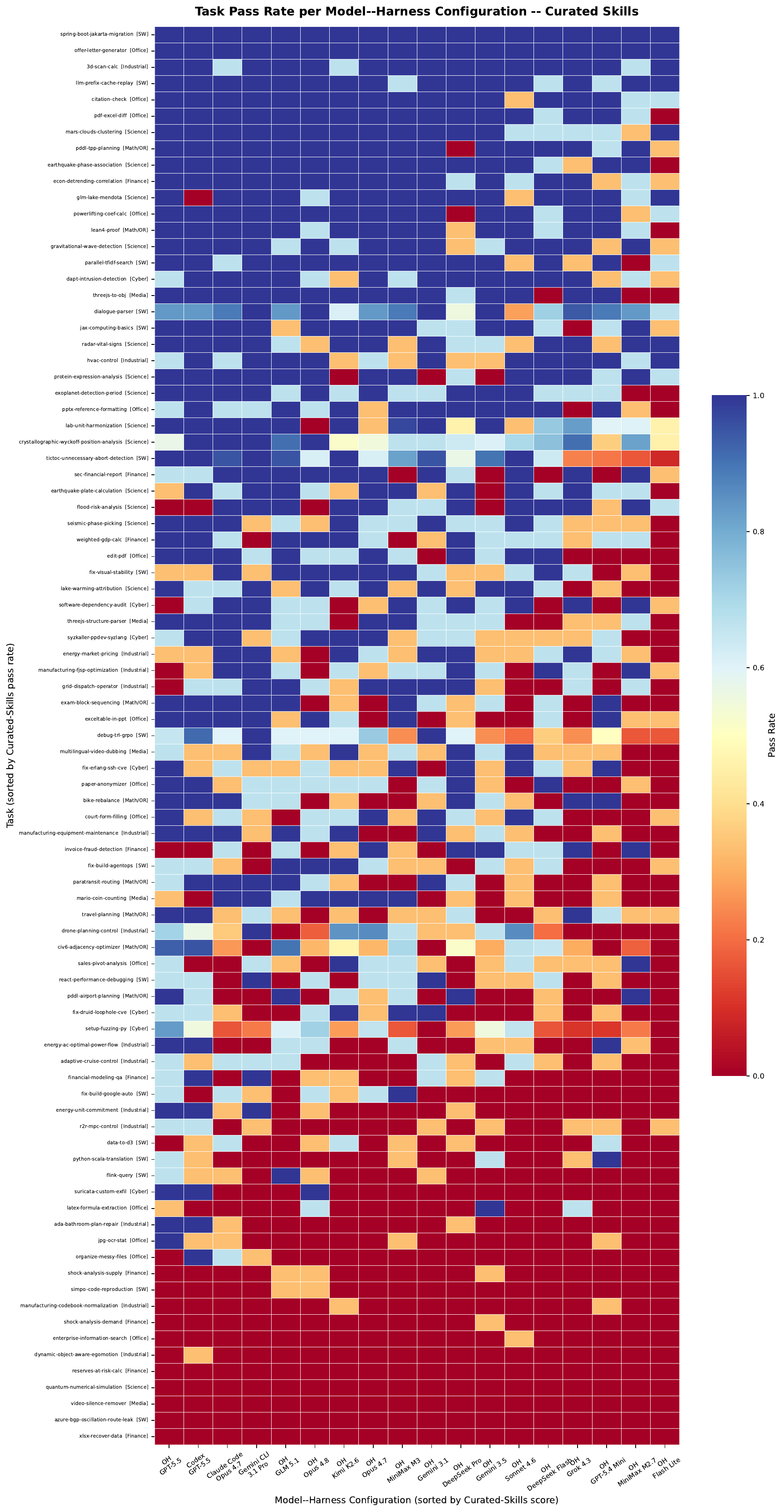}
    \caption{Task pass rate per model with curated Skills. The grid reveals a common set of easy tasks (top, uniformly blue) solved by all models, and hard tasks (bottom, uniformly red) unsolved even with Skills. Tasks along the diagonal transition from solvable to unsolvable as model capability decreases.}
    \label{fig:heatmap-with}
  \end{center}
\end{figure}

\begin{figure}[p]
  \begin{center}
    \includegraphics[width=\textwidth,height=0.88\textheight,keepaspectratio]{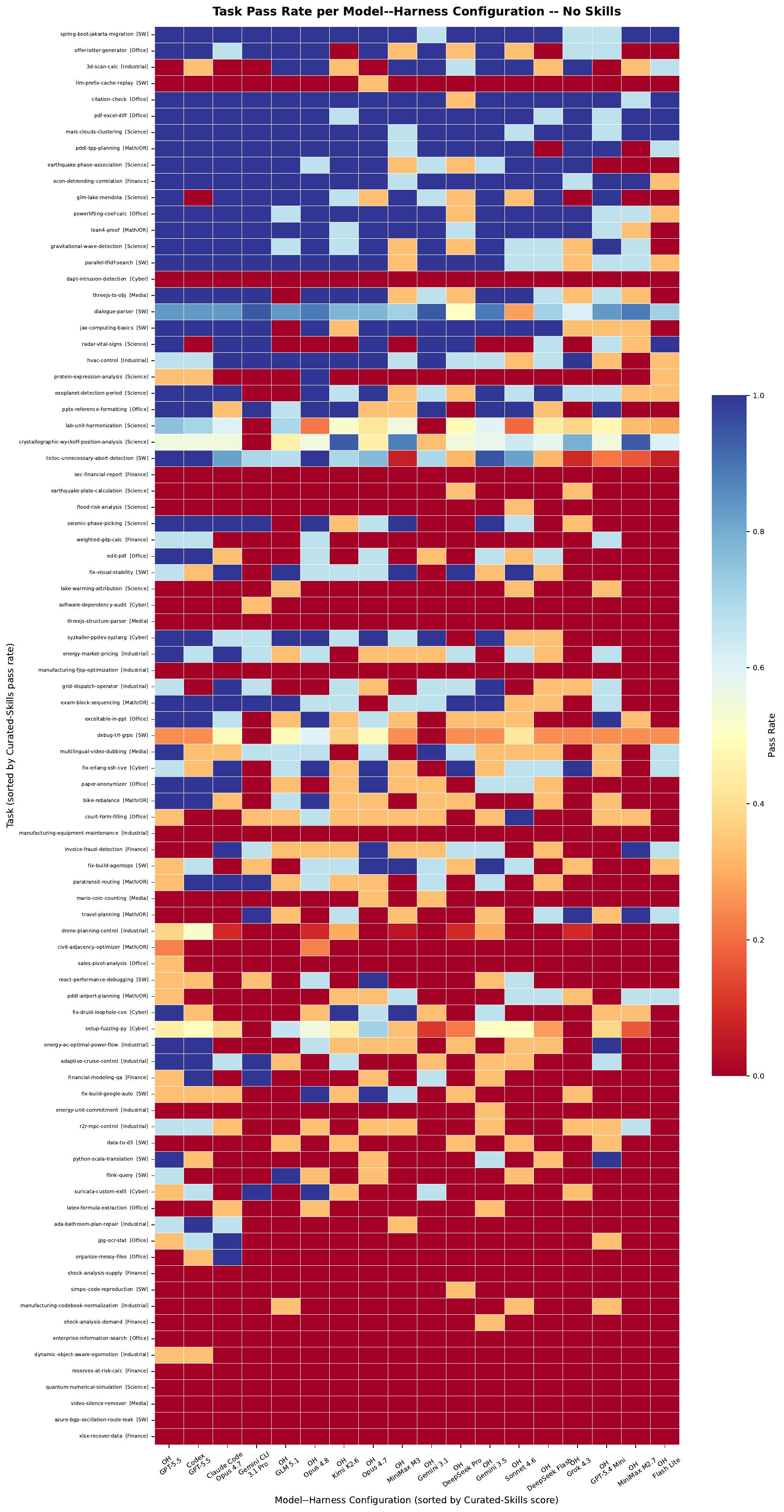}
    \caption{Task pass rate per model without Skills (baseline). Compared to \autoref{fig:heatmap-with}, the blue region contracts substantially, confirming that Skills shift many tasks from unsolved to solved.}
    \label{fig:heatmap-no}
  \end{center}
\end{figure}

\begin{figure}[p]
  \begin{center}
    \includegraphics[width=\textwidth,height=0.88\textheight,keepaspectratio]{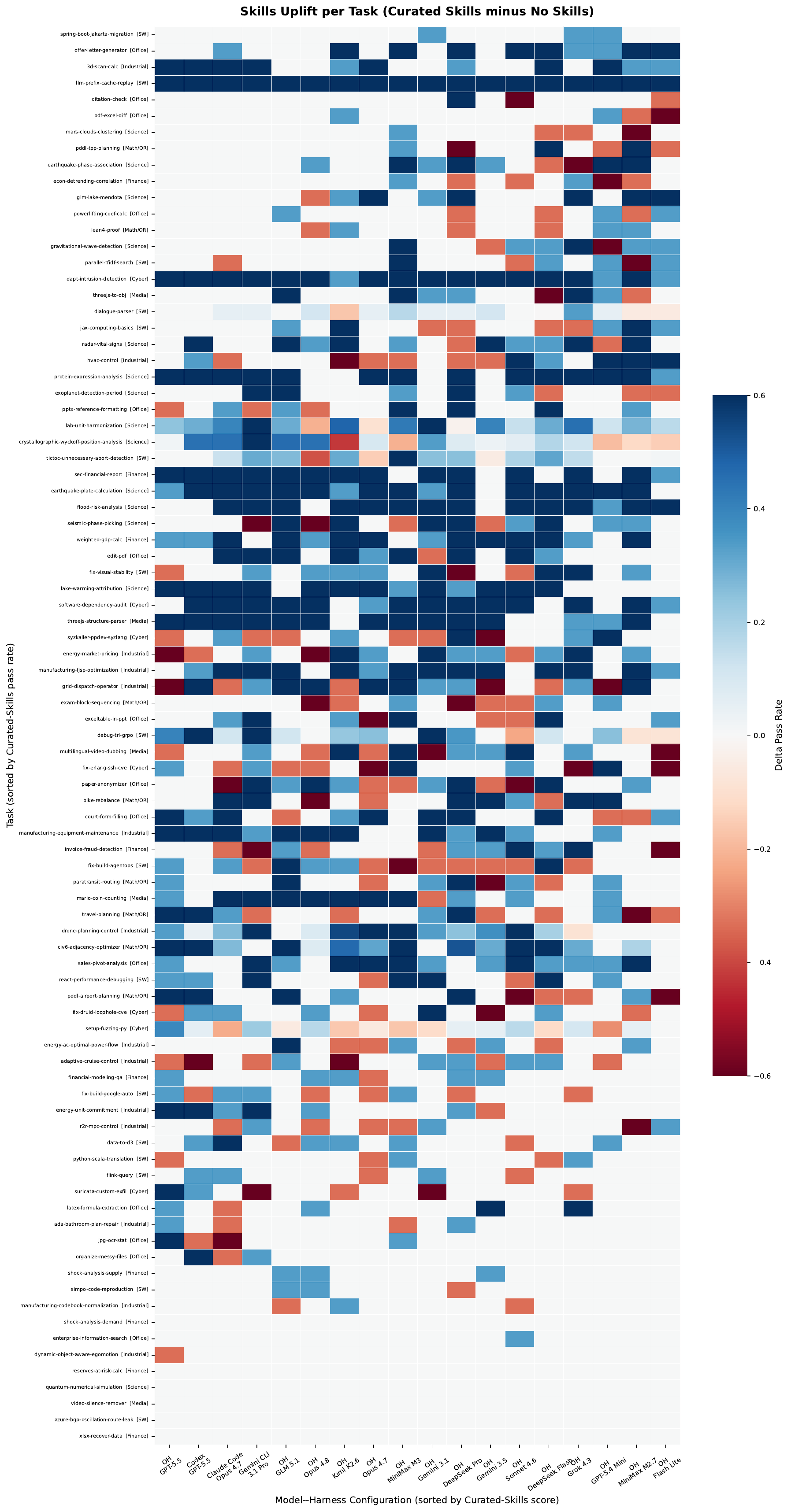}
    \caption{Skills uplift per task (with Skills $-$ without Skills). Blue cells indicate positive uplift; red cells indicate tasks where Skills hurt performance. The majority of cells are blue, confirming broad Skill benefit. A small number of tasks show negative delta for specific models.}
    \label{fig:heatmap-uplift}
  \end{center}
\end{figure}

\autoref{fig:heatmap-with}, \autoref{fig:heatmap-no}, and \autoref{fig:heatmap-uplift} show the task-level pass rates per model across all 87 tasks.
Results use the same fixed three-trial denominator as \autoref{tab:main-results}.
Tasks (rows) are sorted by average with-Skills pass rate; models (columns) are sorted by aggregate with-Skills score.

\section{Skill Design Factors}
\label{app:skill-design-factors}

We expand on the design ablations summarized in Finding~5 (\S\ref{sec:results}). All numbers use the latest 87-task, 18-configuration aggregate with the same fixed three-trial denominator as \autoref{tab:main-results}.

\subsection{Skill Quantity}

\begin{table}[h]
\centering
\small
\caption{Pass rates by number of Skills attached to a task. One to three Skills remain strongest; $\geq 4$ Skills introduces diminishing returns consistent with cognitive overhead and conflicting guidance.}
\label{tab:skill-quantity}
\begin{tabular}{lcccc}
\toprule
\textbf{Skills Count} & \textbf{N tasks} & \textbf{No Skills} & \textbf{With Skills} & \textbf{$\Delta_\text{abs}$} \\
\midrule
\textbf{1} & \textbf{23} & \textbf{32.2\%} & \textbf{50.2\%} & \textbf{\ppos{18.0}} \\
2--3 & 43 & 34.4\% & 53.4\% & \ppos{19.0} \\
$\geq4$ & 21 & 34.8\% & 44.9\% & \ppos{10.1} \\
\bottomrule
\end{tabular}
\end{table}

\subsection{Skill Complexity}

\begin{table}[h]
\centering
\small
\caption{Pass rates by Skill documentation length. Buckets use total \texttt{SKILL.md} text size with 25th, 50th, and 95th percentile cutpoints. Focused documentation outperforms comprehensive prose.}
\label{tab:skill-complexity}
\begin{tabular}{lcccc}
\toprule
\textbf{Complexity} & \textbf{N tasks} & \textbf{No Skills} & \textbf{With Skills} & \textbf{$\Delta_\text{abs}$} \\
\midrule
Compact & 22 & 37.6\% & 56.6\% & \ppos{19.0} \\
Standard & 22 & 19.8\% & 41.2\% & \ppos{21.5} \\
Detailed & 38 & 39.3\% & 53.8\% & \ppos{14.5} \\
Comprehensive & 5 & 39.3\% & 40.0\% & \ppos{0.7} \\
\bottomrule
\end{tabular}
\end{table}

\subsection{Failure Modes when Skills Hurt}
\label{app:skill-failure-modes}

We audited paired (with-Skills, no-Skills) trajectories on tasks with negative deltas. The latest aggregate changes the task ordering, but the same three repeatable patterns remain:

\begin{itemize}[nosep,leftmargin=*]
\item \textbf{Pattern A: heavyweight pipeline crowds out simpler execution.} \emph{Tasks:} \texttt{adaptive-cruise-control} ($-5.6$\,pp), \texttt{r2r-mpc-control} ($-5.6$\,pp), \texttt{dynamic-object-aware-egomotion} ($-1.9$\,pp). The Skill points agents toward more principled but heavier optimization workflows, which can crowd out a simpler path to a valid answer. \emph{Implication:} authors should mark optional steps and supply a fast path.
\item \textbf{Pattern B: Skill activation displaces a stronger native strategy.} \emph{Tasks:} \texttt{exam-block-sequencing} ($-7.4$\,pp), \texttt{econ-detrending-correlation} ($-5.6$\,pp), \texttt{python-scala-translation} ($-1.9$\,pp). A generic recipe can suppress a stronger direct coding or debugging strategy. \emph{Implication:} Skills should include applicability boundaries, not just preferred procedures.
\item \textbf{Pattern C: Skill points the agent at a solver it can't debug.} \emph{Tasks:} \texttt{suricata-custom-exfil} ($-7.4$\,pp), \texttt{mars-clouds-clustering} ($-5.6$\,pp), \texttt{fix-erlang-ssh-cve} ($-3.7$\,pp). When a Skill mandates a brittle framework or schema, agents inherit a steeper debug surface and can fail before the model reaches a correct artifact. \emph{Implication:} Skills should provide debugging fallbacks and validation checks.
\end{itemize}

\subsection{Skill-Author Patterns Audit}
\label{app:skill-author-patterns}

We audited paired trajectories (with-Skills vs.\ no-Skills) on the 10 highest-$\Delta$ tasks (Top-10 in \S\ref{sec:results}, mean $+67.0$\,pp) to extract concrete authoring patterns. The five patterns summarized in §6's ``Implications for Skill authoring'' paragraph are:
(1)~ship an executable script with calibrated defaults rather than only describing the algorithm (\texttt{llm-prefix-cache-replay}, \texttt{sec-financial-report}); (2)~name the canonical data source and parsing quirk (\texttt{flood-risk-analysis}, \texttt{dapt-intrusion-detection}); (3)~encode the exact file-format constraint the verifier inspects (\texttt{threejs-structure-parser}); (4)~surface the algorithmic invariant the verifier asserts (\texttt{protein-expression-analysis}, \texttt{earthquake-plate-calculation}, \texttt{manufacturing-fjsp-optimization}); (5)~make the Skill's \texttt{description:} frontmatter so task-specific that the agent matches it on the first scan. Each pattern is grounded in paired trajectory excerpts, available alongside the released trajectories.

\section{Confidence Interval Calculation}
\label{app:details}


For the main result bars in \autoref{fig:skills_results} and the Skill Invocation Rate plot in \autoref{fig:skill-invocation-rate}, we compute 95\% binomial (Wald) confidence intervals as $1.96\sqrt{p(1-p)/n}$, where $p$ is the corresponding rate and $n$ is the selected public result-file count for that model--harness cell. Resolution-rate intervals are plotted as slightly offset circles within each stacked bar; invocation intervals are plotted as black diamonds with vertical ranges.
Normalized gain is reported as a point estimate.



\section{Complete Task List}
\label{app:task-list}

The complete 87-task inventory and updated domain distribution are reported in \autoref{tab:task-domain-mapping}. The table is generated from \texttt{scripts/taxonomy.csv} and grouped by the updated 8-domain taxonomy.


\section{Comprehensive Results Summary}
\label{app:results-summary}

\autoref{tab:full-results} consolidates the latest aggregate results across 18 model--harness configurations and two matched Skills conditions. Scores are computed on the fixed $87\times3$ trial frame per condition, consistent with the task-macro scoring used in the main text; the $n$ column reports current public coverage.

\vspace{-8pt}

\begin{table}[htb]
\centering
\footnotesize
\caption{Comprehensive latest results across model--harness configurations. Pass rates use a fixed $87\times3$ denominator per condition; $\Delta$ is curated-Skills improvement and $g$ is normalized gain. Models are ordered by with-Skills pass rate; $n$ reports selected no-Skills / curated-Skills result files.}
\label{tab:full-results}
\setlength{\tabcolsep}{0pt}
\renewcommand{\arraystretch}{1.05}
\begin{tabular*}{0.94\textwidth}{@{\extracolsep{\fill}}llrrrrr@{}}
\toprule
\textbf{Harness} & \textbf{Model} & \textbf{No Sk.} & \textbf{With Sk.} & \textbf{$\Delta$ (pp)} & \textbf{$g$ (\%)} & \textbf{$n$} \\
\midrule
OpenHands & GPT-5.5 & 51.5 & 67.3 & \ppos{15.8} & 32.6 & 261/261 \\
Codex & GPT-5.5 & 46.8 & 66.5 & \ppos{19.7} & 37.0 & 261/261 \\
Claude Code & Opus 4.7 & 43.0 & 61.2 & \ppos{18.2} & 31.9 & 261/261 \\
Gemini CLI & Gemini 3.1 Pro & 36.0 & 60.8 & \ppos{24.8} & 38.7 & 261/261 \\
OpenHands & GLM 5.1 & 32.7 & 58.4 & \ppos{25.7} & 38.1 & 261/261 \\
OpenHands & Claude Opus 4.8 & 45.7 & 54.1 & \ppos{8.4} & 15.5 & 261/261 \\
OpenHands & Kimi K2.6 & 33.4 & 54.0 & \ppos{20.6} & 31.0 & 261/261 \\
OpenHands & Claude Opus 4.7 & 42.1 & 53.1 & \ppos{11.1} & 19.1 & 261/261 \\
OpenHands & MiniMax M3 & 29.7 & 53.0 & \ppos{23.3} & 33.2 & 261/261 \\
OpenHands & Gemini 3.1 Pro & 33.8 & 52.8 & \ppos{19.0} & 28.7 & 261/261 \\
OpenHands & DeepSeek V4 Pro & 26.9 & 50.1 & \ppos{23.2} & 31.8 & 261/261 \\
OpenHands & Gemini 3.5 Flash & 41.1 & 48.2 & \ppos{7.1} & 12.1 & 261/261 \\
OpenHands & Claude Sonnet 4.6 & 33.5 & 47.2 & \ppos{13.6} & 20.5 & 261/261 \\
OpenHands & DeepSeek V4 Flash & 27.5 & 44.7 & \ppos{17.2} & 23.7 & 261/261 \\
OpenHands & Grok 4.3 & 22.8 & 41.7 & \ppos{18.8} & 24.4 & 261/261 \\
OpenHands & GPT-5.4 Mini & 29.9 & 41.4 & \ppos{11.5} & 16.4 & 261/261 \\
OpenHands & MiniMax M2.7 & 18.1 & 34.9 & \ppos{16.8} & 20.5 & 261/261 \\
OpenHands & Gemini 3.1 Flash Lite & 16.0 & 20.1 & \ppos{4.1} & 4.9 & 261/261 \\
\midrule
\textbf{Mean} & & \textbf{33.9} & \textbf{50.5} & \textbf{\ppos{16.6}} & \textbf{25.5} & \textbf{4698/4698} \\
\bottomrule
\end{tabular*}
\end{table}

\paragraph{Key observations.}
\begin{itemize}[nosep]
\item Curated Skills improve performance by +16.6\,pp on average (range: +4.1 to +25.7\,pp), corresponding to a normalized gain of 25.5\%.
\item OpenHands + GPT-5.5 achieves the highest absolute pass rate (67.3\%) with Skills.
\item OpenHands + GLM~5.1 shows the largest absolute improvement (+25.7\,pp), while Gemini CLI + Gemini~3.1 Pro has the highest normalized gain (38.7\%).
\item The current public snapshot is complete for all 18 model--harness configurations.
\end{itemize}

\begin{figure}[htb]
\centering
\includegraphics[width=\textwidth]{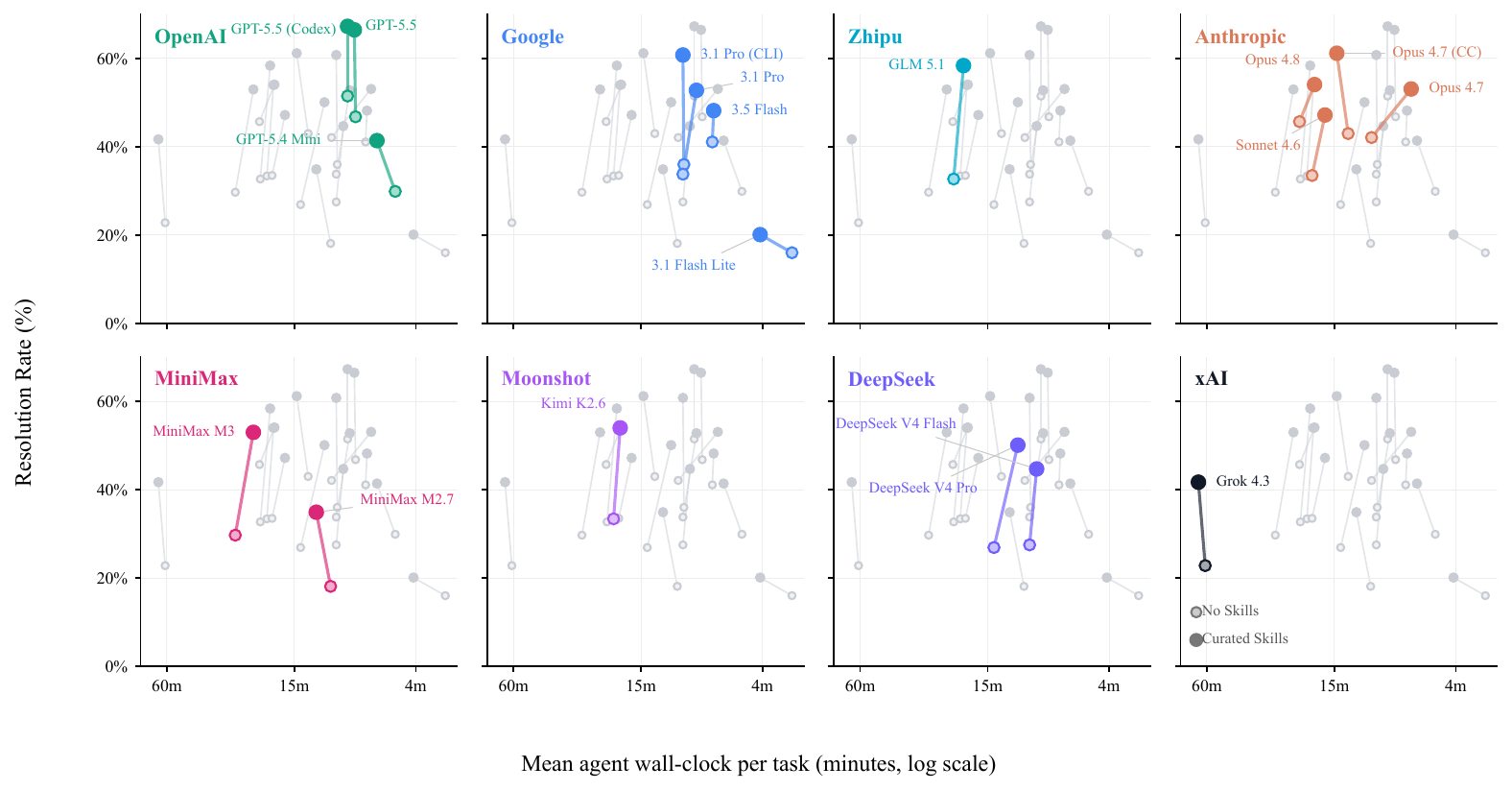}
\caption{Per-family shift from no Skills (hollow) to curated Skills (solid) in the time--performance plane; gray shows the remaining fleet.}
\label{fig:time-vs-score-vendor}
\end{figure}

\autoref{fig:time-vs-score-vendor} breaks the same aggregate down by model family in the time--performance plane (mean agent wall-clock per task; measurement details in \appautoref{app:time-performance}): every family improves with curated Skills, none at a material time cost.

\subsection{Auxiliary Failure-Mode Snapshot}

\vspace{-8pt}

\label{app:failure-mode-snapshot}

\begin{table}[htb]
\centering
\small
\caption{Failure mode distribution by condition in the latest healthy-first diagnostic selection. Fail Rate uses the fixed 4{,}698-slot condition denominator; mode shares are percentages of that condition's selected non-solved trials. The selected public coverage spans the full 9{,}396-slot frame.}
\label{tab:failure-mode-snapshot}
\begin{tabular}{lrrrrr}
\toprule
\textbf{Condition} & \textbf{Fail Rate} & \textbf{Timeout} & \textbf{Execution} & \textbf{Coherence} & \textbf{Verification} \\
\midrule
No Skills & 66.1\% & 1.9\% & 0.0\% & 7.9\% & 90.3\% \\
With Skills & 49.5\% & 3.1\% & 0.0\% & 9.1\% & 87.8\% \\
\bottomrule
\end{tabular}
\end{table}

\begin{table}[htb]
\centering
\small
\caption{Outcome distribution by configuration (both Skills conditions pooled; \% of each configuration's selected trials). Ordered by solved rate.}
\label{tab:outcome-by-config}
\begin{tabular*}{\textwidth}{@{\extracolsep{\fill}}llrrrr@{}}
\toprule
\textbf{Harness} & \textbf{Model} & \textbf{Solved} & \textbf{Partial} & \textbf{Attempted} & \textbf{Runtime err.} \\
\midrule
OpenHands & GPT-5.5 & 55.2\% & 6.5\% & 38.3\% & 0.0\% \\
Codex & GPT-5.5 & 52.9\% & 5.6\% & 41.6\% & 0.0\% \\
Claude Code & Opus 4.7 & 49.0\% & 5.4\% & 45.6\% & 0.0\% \\
Gemini CLI & Gemini 3.1 Pro & 48.1\% & 0.6\% & 51.3\% & 0.0\% \\
OpenHands & Claude Opus 4.8 & 47.3\% & 4.2\% & 48.5\% & 0.0\% \\
OpenHands & Claude Opus 4.7 & 43.9\% & 5.7\% & 50.4\% & 0.0\% \\
OpenHands & Gemini 3.1 Pro & 42.9\% & 0.8\% & 56.3\% & 0.0\% \\
OpenHands & Gemini 3.5 Flash & 41.6\% & 5.6\% & 52.9\% & 0.0\% \\
OpenHands & GLM 5.1 & 41.4\% & 6.1\% & 52.5\% & 0.0\% \\
OpenHands & Kimi K2.6 & 40.6\% & 5.6\% & 51.9\% & 1.9\% \\
OpenHands & MiniMax M3 & 38.9\% & 5.4\% & 55.7\% & 0.0\% \\
OpenHands & Claude Sonnet 4.6 & 38.3\% & 4.2\% & 47.7\% & 9.8\% \\
OpenHands & DeepSeek V4 Pro & 35.4\% & 6.5\% & 56.7\% & 1.3\% \\
OpenHands & GPT-5.4 Mini & 33.0\% & 5.7\% & 61.3\% & 0.0\% \\
OpenHands & DeepSeek V4 Flash & 32.8\% & 6.7\% & 60.0\% & 0.6\% \\
OpenHands & Grok 4.3 & 30.1\% & 5.7\% & 64.2\% & 0.0\% \\
OpenHands & MiniMax M2.7 & 23.9\% & 5.6\% & 69.2\% & 1.3\% \\
OpenHands & Gemini 3.1 Flash Lite & 15.9\% & 5.6\% & 78.5\% & 0.0\% \\
\bottomrule
\end{tabular*}
\end{table}

\vspace{-8pt}

\section{Task-Level Case Studies}
\label{app:aux-diagnostics}

The following task-level audits use verifier-scored pass/fail outcomes to illustrate how curated Skills change agent behavior. They are qualitative supplements to the aggregate pass-rate tables above.

\subsection{Success Case Studies}
\label{app:success-cases}

We present representative examples where curated Skills transformed agent outcomes from failure to success, illustrating the mechanisms through which procedural knowledge improves performance.

\paragraph{Skills bridge domain-specific API gaps: \texttt{sales-pivot-analysis}.}
Without Skills, evaluated agents rarely solved this task, which requires creating Excel pivot tables programmatically from population and income data. Agents consistently loaded the data correctly but failed at pivot table creation---Codex attempted manual DataFrame reshaping instead of using \texttt{openpyxl}'s pivot table API, producing structurally incorrect output (10/23 tests failed with ``list index out of range'' on missing pivot objects). With Skills providing step-by-step guidance for the \texttt{openpyxl} pivot table workflow, average pass rate rises from 1.9\% to 40.7\% (+38.9\,pp).

\paragraph{Skills provide critical data processing pipelines: \texttt{flood-risk-analysis}.}
This task requires identifying flood-risk stations from USGS streamflow data using return period estimation. Without Skills, agents attempted ad-hoc statistical approaches---e.g., simple threshold-based detection or incorrect distribution fitting---achieving only 1.9\% pass rate. The curated Skill specified the Log-Pearson Type III distribution, the standard USGS methodology for flood frequency analysis, including the exact scipy function calls and parameter interpretation. With Skills, pass rate rose to 68.5\% (+66.7\,pp), with many configurations correctly applying the USGS-standard methodology.

\paragraph{Skills encode regulatory knowledge: \texttt{sec-financial-report}.}
Analyzing hedge fund activities from SEC 13F filings requires understanding specific regulatory formats, CIK lookup procedures, and filing comparison methodology. Without Skills, no model could complete the task (0\% pass rate)---agents either failed to locate the correct filings or misinterpreted the tabular data format. The curated Skill documented the SEC EDGAR API endpoints, 13F-HR filing structure, and cross-quarter comparison methodology. With Skills, pass rate reached 68.5\% (+68.5\,pp).

\paragraph{Skills prevent common implementation pitfalls: \texttt{manufacturing-fjsp-optimization}.}
The flexible job-shop scheduling problem requires constraint-aware optimization with machine downtime windows. Without Skills, agents produced naive schedules ignoring maintenance constraints (0\% pass rate). The curated Skill outlined the constraint propagation approach, objective function formulation, and OR-Tools solver configuration. With Skills, agents successfully formulated and solved the optimization problem in a subset of configurations (55.6\% pass rate, +55.6\,pp).

\vspace{-8pt}

\subsection{Qualitative Failure Examples}
\label{app:failure-examples}

We present representative examples drawn from verifier test outputs across failed trajectories.

\paragraph{Quality Below Threshold: \texttt{earthquake-plate-calculation}.}
The agent correctly identified the target earthquake event, extracting latitude, longitude, magnitude, and timestamp accurately (7/8 tests passed). However, the computed distance from the nearest plate boundary was 3,562 km instead of the expected 3,878 km---an 8.2\% error that exceeded the $\pm$0.01 km tolerance. The agent applied the correct Haversine formula but used an incorrect plate boundary coordinate, demonstrating that even when agents understand the computational method, domain-specific data interpretation remains error-prone.

\vspace{-8pt}

\paragraph{Incomplete Solution: \texttt{shock-analysis-supply}.}
The agent created a structurally correct Excel workbook and passed 6 of 9 tests, correctly setting up sheet structures, formula templates, and some data imports. However, it failed to: (1) populate employment/labor data from the Penn World Tables (PWT), (2) execute the HP filter optimization solver, and (3) compute the depreciation rate. These three missing components represent the most domain-specific and computationally demanding aspects of the task.

\paragraph{No Output Produced: \texttt{gh-repo-analytics}.}
All 8 verifier tests errored at the fixture stage with ``Missing \texttt{/app/report.json},'' meaning the agent never created the required output file. The task requires interacting with a local Gitea server, cloning repositories, and computing analytics---a multi-step pipeline where failure at any early stage prevents all downstream output.

\paragraph{Specification Violation: \texttt{latex-formula-extraction}.}
The agent extracted LaTeX formulas from a PDF but included markdown headers alongside the formulas in the output file, producing 6 entries instead of the required 5. The specification required one formula per line wrapped in \texttt{\$\$} delimiters; the extraneous headers violated this format constraint. This failure illustrates agents' tendency to over-include content rather than strictly adhering to output specifications.

\paragraph{Domain Knowledge Gap: \texttt{exceltable-in-ppt}.}
The agent correctly updated the primary exchange rate cell in the PowerPoint-embedded Excel table (6/8 tests passed) but failed to recompute inverse rates and dependent cells, producing \texttt{NaN} propagation through the spreadsheet. The underlying issue was misunderstanding how Excel formula dependencies cascade in embedded workbooks---a domain-specific detail that Skills could address.

\paragraph{Skills transforming outcomes: \texttt{sales-pivot-analysis}.}
Without Skills, Codex populated source data correctly but could not create Excel pivot tables (10/23 tests failed with ``list index out of range'' on missing pivot objects). With Skills, the Skills provided Office-specific guidance for programmatic pivot table creation that the agent could not discover independently. In the latest aggregate, this task improves from 1.9\% to 40.7\% (+38.9\,pp), illustrating how Skills can bridge a specific capability gap.




\subsection{Tasks With Largest Skills Impact}
\label{app:skills-delta}

\autoref{tab:skills-delta} lists the 10 tasks where Skills produced the largest improvement in pass rate. These tasks share a common pattern: they require domain-specific procedural knowledge (e.g., cache behavior, intrusion detection, financial reporting schemas, scientific data processing pipelines, optimization, and artifact conversion) that is well-suited to being encoded in Skill documents. The average improvement for these top-10 tasks is +67.0 percentage points.

\begin{table}[htb]
\centering
\small
\caption{Tasks with largest Skills impact (With Skills pass rate $-$ No Skills pass rate), using the latest fixed three-trial denominator.}
\label{tab:skills-delta}
\begin{tabular}{lccc}
\toprule
\textbf{Task} & \textbf{No Skills} & \textbf{With Skills} & \textbf{$\Delta$} \\
\midrule
llm-prefix-cache-replay & 1.9\% & 94.4\% & +92.6\,pp \\
dapt-intrusion-detection & 0.0\% & 81.5\% & +81.5\,pp \\
sec-financial-report & 0.0\% & 68.5\% & +68.5\,pp \\
flood-risk-analysis & 1.9\% & 68.5\% & +66.7\,pp \\
protein-expression-analysis & 11.1\% & 77.8\% & +66.7\,pp \\
earthquake-plate-calculation & 3.7\% & 68.5\% & +64.8\,pp \\
software-dependency-audit & 1.9\% & 61.1\% & +59.3\,pp \\
threejs-structure-parser & 0.0\% & 59.3\% & +59.3\,pp \\
lake-warming-attribution & 5.6\% & 61.1\% & +55.6\,pp \\
manufacturing-fjsp-optimization & 0.0\% & 55.6\% & +55.6\,pp \\
\bottomrule
\end{tabular}
\end{table}

\vspace{-8pt}

\section{Skill Invocation Rate}
\label{app:skill-invocation-rate}

\autoref{tab:skill-invocation-rate} reports the exact counts behind \autoref{fig:skill-invocation-rate}. Counts include only task-specific Skills shipped under each task's \texttt{environment/skills}; harness-bundled Skills are treated as part of the harness.

\begin{table}[htb]
\centering
\footnotesize
\caption{Skill Invocation Rate by model--harness configuration on curated-Skills trials. Counts require a task-bundled Skill; harness-bundled Skills are excluded.}
\label{tab:skill-invocation-rate}
\setlength{\tabcolsep}{4pt}
\begin{tabular*}{\textwidth}{@{\extracolsep{\fill}}llrrrr@{}}
\toprule
\textbf{Harness} & \textbf{Model} & \textbf{Skill Invocation Rate} & \textbf{Invoked} & \textbf{Resolution Rate} & \textbf{Resolution $\mid$ Invoked} \\
\midrule
Codex & GPT-5.5 & 99.2\% & 259/261 & 66.5\% & 66.2\% \\
OpenHands & GPT-5.5 & 92.0\% & 240/261 & 67.3\% & 68.4\% \\
Gemini CLI & Gemini 3.1 Pro & 90.4\% & 236/261 & 60.8\% & 63.0\% \\
OpenHands & Claude Sonnet 4.6 & 88.9\% & 232/261 & 47.2\% & 49.2\% \\
OpenHands & Grok 4.3 & 84.7\% & 221/261 & 41.7\% & 46.1\% \\
OpenHands & Claude Opus 4.8 & 83.9\% & 219/261 & 54.1\% & 54.5\% \\
OpenHands & GLM 5.1 & 76.2\% & 199/261 & 58.4\% & 61.7\% \\
OpenHands & Claude Opus 4.7 & 74.3\% & 194/261 & 53.1\% & 54.7\% \\
OpenHands & MiniMax M2.7 & 70.5\% & 184/261 & 34.9\% & 40.9\% \\
OpenHands & DeepSeek V4 Pro & 69.3\% & 181/261 & 50.1\% & 56.8\% \\
OpenHands & MiniMax M3 & 68.6\% & 177/258 & 53.0\% & 54.6\% \\
Claude Code & Opus 4.7 & 68.2\% & 178/261 & 61.2\% & 63.9\% \\
OpenHands & DeepSeek V4 Flash & 64.4\% & 168/261 & 44.7\% & 49.8\% \\
OpenHands & GPT-5.4 Mini & 62.5\% & 163/261 & 41.4\% & 45.6\% \\
OpenHands & Kimi K2.6 & 57.9\% & 151/261 & 54.0\% & 57.3\% \\
OpenHands & Gemini 3.5 Flash & 55.9\% & 146/261 & 48.2\% & 51.0\% \\
OpenHands & Gemini 3.1 Pro & 54.0\% & 141/261 & 52.8\% & 68.7\% \\
OpenHands & Gemini 3.1 Flash Lite & 46.4\% & 121/261 & 20.1\% & 29.8\% \\
\bottomrule
\end{tabular*}
\end{table}

\section{Token, Cost, and Time Efficiency}
\label{app:token-cost}

This section reports token usage, per-trial cost, and agent wall-clock time for the latest 18-configuration aggregate, computed directly from the selected public \texttt{result.json} files (HF \texttt{main}\,+\,PR\#11 fixed at \texttt{a26b2100}, across \texttt{v0.1} and \texttt{v1.1} submissions) over the same healthy-first three-trial selection used in the main result tables. Each trial records its own token counts and agent-execution timing; where the provider is priced through our LiteLLM runtime it also records a per-trial cost in USD. Token counts are available for all 18 configurations and LiteLLM cost for 10 of them (several open-weight and CLI providers are not yet priced in our runtime). We average over selected trials with recorded usage or cost and suppress any cell backed by fewer than 20 usable trials.

\subsection{Token Usage and Cost by Configuration}

\autoref{tab:token-usage} reports mean total tokens per trial (prompt\,+\,completion, as recorded by the harness) and mean LiteLLM cost per trial, separately for the no-Skills and curated-Skills conditions.

\begin{table}[htb]
\centering
\footnotesize
\caption{Mean per-trial token usage and cost in the latest 18-configuration aggregate, by Skills condition. \textbf{Tok} = mean total tokens per trial (thousands); \textbf{\$} = mean LiteLLM-computed cost per trial. Cells with fewer than 20 usable trials are shown as ``--''; ``--'' in a \$ column also marks a provider not priced through our LiteLLM runtime. Claude Code counts are session-JSONL-derived and approximate.}
\label{tab:token-usage}
\setlength{\tabcolsep}{5pt}
\begin{tabular}{ll rr rr}
\toprule
& & \multicolumn{2}{c}{\textbf{No Skills}} & \multicolumn{2}{c}{\textbf{Curated Skills}} \\
\cmidrule(lr){3-4}\cmidrule(lr){5-6}
\textbf{Harness} & \textbf{Model} & \textbf{Tok (K)} & \textbf{\$} & \textbf{Tok (K)} & \textbf{\$} \\
\midrule
OpenHands & GPT-5.5 & 1{,}197 & 1.73 & 1{,}267 & 1.67 \\
Gemini CLI & Gemini 3.1 Pro & 1{,}165 & -- & 1{,}932 & -- \\
OpenHands & GLM 5.1 & 2{,}371 & -- & 2{,}117 & -- \\
Codex & GPT-5.5 & 3{,}222 & 3.08 & 3{,}227 & 2.99 \\
Claude Code & Opus 4.7$^\dagger$ & 4{,}332 & 5.21 & 6{,}425 & 6.74 \\
OpenHands & Claude Opus 4.7 & 1{,}771 & 10.99 & 1{,}094 & 6.37 \\
OpenHands & Claude Opus 4.8 & 2{,}914 & 22.70 & 2{,}432 & 14.02 \\
OpenHands & Gemini 3.1 Pro & 1{,}818 & 0.84 & 4{,}018 & 1.88 \\
OpenHands & MiniMax M3 & 4{,}718 & -- & 4{,}404 & -- \\
OpenHands & Gemini 3.5 Flash & 2{,}827 & 1.24 & 2{,}675 & 1.18 \\
OpenHands & Kimi K2.6 & 2{,}157 & -- & 2{,}103 & -- \\
OpenHands & DeepSeek V4 Pro & 1{,}832 & -- & 1{,}687 & -- \\
OpenHands & Claude Sonnet 4.6 & 1{,}834 & 12.34 & 2{,}874 & 10.15 \\
OpenHands & Grok 4.3 & 489 & -- & 579 & -- \\
OpenHands & DeepSeek V4 Flash & 1{,}923 & -- & 2{,}449 & -- \\
OpenHands & MiniMax M2.7 & 1{,}517 & -- & 2{,}969 & -- \\
OpenHands & GPT-5.4 Mini & 1{,}282 & 0.25 & 1{,}393 & 0.30 \\
OpenHands & Gemini 3.1 Flash Lite & 3{,}186 & 0.15 & 4{,}828 & 0.22 \\
\bottomrule
\end{tabular}

\vspace{2pt}
\footnotesize $^\dagger$Claude Code token counts are derived from session JSONL logs and are approximate (effective input includes cache-creation and cache-read tokens).
\end{table}

\subsection{Outcome and Runtime Diagnostics}
\label{app:outcome-runtime}

\begin{table}[htb]
\centering
\small
\caption{Trial-level outcome distribution in the latest 18-configuration aggregate ($N=9{,}396$ selected public trials), overall and by Skills condition. \textbf{Attempted} = a verifier-scored trial with reward 0; \textbf{Runtime err.} = no scored result (agent/harness/verifier error), counted as reward 0 in the main aggregate.}
\label{tab:outcome-distribution}
\begin{tabular}{lrrrrr}
\toprule
\textbf{Condition} & \textbf{$n$} & \textbf{Solved} & \textbf{Partial} & \textbf{Attempted} & \textbf{Runtime err.} \\
\midrule
No Skills & 4{,}698 & 31.3\% & 5.4\% & 62.7\% & 0.6\% \\
Curated Skills & 4{,}698 & 47.7\% & 4.7\% & 46.5\% & 1.0\% \\
\midrule
\textbf{Overall} & \textbf{9{,}396} & \textbf{39.5\%} & \textbf{5.1\%} & \textbf{54.6\%} & \textbf{0.8\%} \\
\bottomrule
\end{tabular}
\end{table}

\begin{table}[htb]
\centering
\small
\caption{Runtime errors in the latest 18-configuration aggregate (trials with no scored result; 78 of 9{,}396 = 0.8\%). Classified from the recorded \texttt{error}/\texttt{error\_category} fields.}
\label{tab:runtime-errors}
\setlength{\tabcolsep}{6pt}
\begin{tabular}{lrl}
\toprule
\textbf{Error class} & \textbf{Count} & \textbf{Source} \\
\midrule
Command timeout & 72 & Shell command exceeded its per-command limit \\
Agent-process crash & 0 & ACP internal/transport error \\
Agent timeout (wall-clock/idle) & 5 & Agent exceeded its wall-clock or idle budget \\
Subprocess crash & 0 & Local subprocess/transport closed unexpectedly \\
Verifier error & 1 & Verifier crash or timeout \\
\midrule
\textbf{Total} & \textbf{78} & 0.8\% of selected public trials \\
\bottomrule
\end{tabular}
\end{table}

\subsection{Cost-Performance Tradeoff}

Per-trial cost spans roughly two orders of magnitude across the lineup. The cheapest priced configurations are OpenHands\,+\,Gemini 3.1 Flash Lite (\$0.15--0.22) and OpenHands\,+\,GPT-5.4 Mini (\$0.25--0.30); the Gemini and GPT-5.5 configurations occupy the mid-range (\$0.8--3.1 per trial); and the Claude Opus configurations are by far the most expensive---OpenHands\,+\,Claude Opus 4.8 at \$14.02--22.70 per trial, OpenHands\,+\,Claude Opus 4.7 at \$6.37--10.99, and OpenHands\,+\,Claude Sonnet 4.6 at roughly \$10--12.3. Cost does not track capability monotonically: OpenHands\,+\,GPT-5.5 attains the highest with-Skills pass rate (67.3\%) at \$1.67 per trial---about eight times cheaper than OpenHands\,+\,Claude Opus 4.8 (54.1\% at \$14.02).

Skills do not move token usage in a single direction. Several configurations consume \emph{more} tokens with Skills (OpenHands\,+\,Gemini 3.1 Pro 1.82M\,$\rightarrow$\,4.02M; OpenHands\,+\,Claude Sonnet 4.6 1.83M\,$\rightarrow$\,2.87M; OpenHands\,+\,Gemini 3.1 Flash Lite 3.19M\,$\rightarrow$\,4.83M), consistent with the agent reading and acting on the additional Skill context, while others consume \emph{fewer} (OpenHands\,+\,Claude Opus 4.7 1.77M\,$\rightarrow$\,1.09M; OpenHands\,+\,Claude Opus 4.8 2.91M\,$\rightarrow$\,2.43M; OpenHands\,+\,Gemini 3.5 Flash 2.83M\,$\rightarrow$\,2.67M), consistent with Skills letting the model reach a solution with less exploratory work. Where providers expose prompt-caching fields, cache reads can account for a material share of effective input tokens, so realized cost under cached pricing can differ substantially from standard-rate estimates.

\subsection{Time--Performance Tradeoff}
\label{app:time-performance}

\begin{figure}[htb]
\centering
\includegraphics[width=\textwidth]{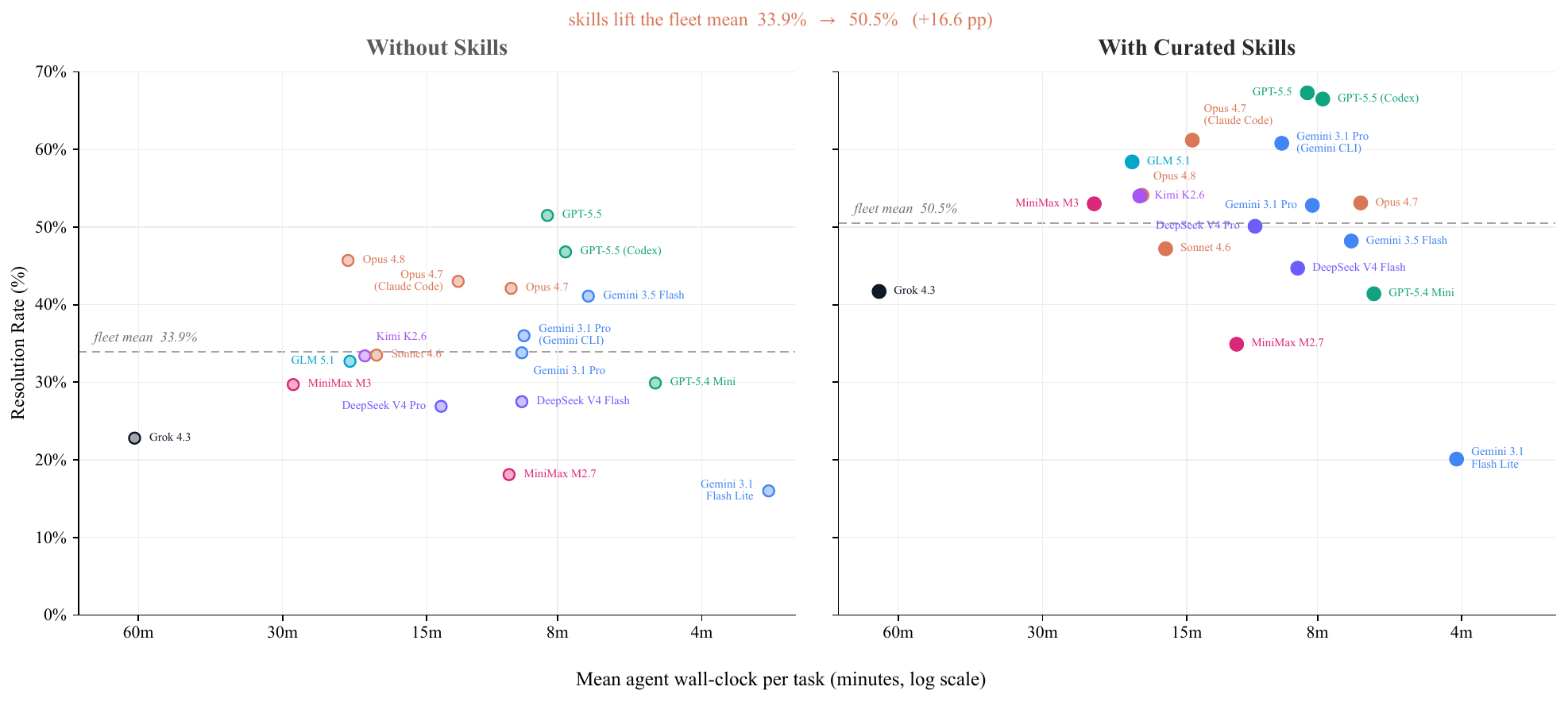}
\caption{Resolution rate vs.\ mean agent wall-clock per task (minutes, log scale), without (left) and with (right) curated Skills; dashed lines mark fleet means.}
\label{fig:time-vs-score}
\end{figure}

\autoref{fig:time-vs-score} plots resolution rate against mean agent wall-clock per task---the agent-execution span of each selected trial (agent runtime only), averaged within each task and then across the 87 tasks---for both Skills conditions, over the same selection as \autoref{tab:full-results}. Curated Skills lift the fleet mean by +16.6\,pp at broadly unchanged agent time; the slower half of the fleet runs faster with Skills (e.g., OpenHands\,+\,Claude Opus 4.7: 10.0\,$\rightarrow$\,6.5 minutes per task). Speed and capability are not in tension: the 4--9-minute band contains both the strongest and the weakest configuration, while the slowest, OpenHands\,+\,Grok 4.3, remains below the with-Skills fleet mean.

\section{\benchmarkName{} Construction Details}
\label{app:skillsbench-construction-details}

This appendix collects the construction and quality-control details summarized in \S\ref{sec:SkillsBench}. It is kept separate from the other appendices so that the benchmark definition, submission process, and leakage safeguards can be inspected as a single unit; the contributor-facing format and PR-level review mechanics (CI commands, checklists, report template, lifecycle) are documented in \appautoref{app:task-spec}.

\subsection{Skill Definition and Boundary}

A \textbf{Skill} is an artifact satisfying four operational criteria:
\begin{itemize}[nosep,leftmargin=*]
    \item \textbf{Procedural content}: Contains how-to guidance, workflows, standard operating procedures, or domain conventions rather than factual retrieval.
    \item \textbf{Task-class applicability}: Applies to a class of problems rather than a single benchmark instance.
    \item \textbf{Structured components}: Includes a \texttt{SKILL.md} file plus optional resources such as scripts, templates, references, or worked examples.
    \item \textbf{Portability}: Is represented as ordinary file-system content, making it easy to edit, version, share, and use across Skills-compatible agent harnesses.
\end{itemize}

\noindent This definition explicitly excludes system prompts, which lack file-structured resources; few-shot examples~\citep{brown2020languagemodelsfewshotlearners}, which are primarily declarative rather than procedural; RAG retrievals~\citep{lewis2021retrievalaugmentedgenerationknowledgeintensivenlp}, which provide factual context rather than procedural workflows; and tool documentation~\citep{schick2023toolformerlanguagemodelsteach, qin2024toolllm}, which describes tool capabilities rather than how to solve a task class. The boundary is not absolute--for example, a StackOverflow answer may mix factual and procedural content--but these criteria provide an operational definition for benchmark construction.

In \benchmarkName{}, each Skill is a modular package under \texttt{environment/skills/}. The required \texttt{SKILL.md} specifies how to approach a task class, while optional resources may include executable scripts, code templates, reference documentation, or examples that the agent can invoke or consult.

\subsection{Task Package}

Each task is a self-contained module with four required components:
\begin{itemize}[nosep,leftmargin=*]
\item \textbf{Instruction.} A human-readable task description specifying the objective, input format, constraints, and expected output. Instructions are written to be solvable by a knowledgeable human without access to the paired Skills, though Skills may substantially reduce time-to-solution.
\item \textbf{Environment.} A Docker container with task-specific data files and a \texttt{skills/} subdirectory containing modular Skills packages. Containerization provides isolated dependencies and clean file-system state.
\item \textbf{Oracle.} A reference solution demonstrating that the task is resolvable. The oracle must pass the verifier with 100\% success before a task can be accepted. (The on-disk directory is \texttt{oracle/}.)
\item \textbf{Verifier.} Deterministic test scripts with programmatic assertions, including numeric tolerances where appropriate. This yields reproducible pass/fail judgments without LLM-as-a-judge variance. (The on-disk directory is \texttt{verifier/}.)
\end{itemize}

\subsection{Community Submission Model}

To maximize domain and workflow diversity, we used a community-driven, open-source contribution model. In total, 142 contributors from academia and industry submitted 400 candidate tasks. We counted submissions that included the full task specification--instruction, environment, oracle, and verifier--along with contributor-assessed difficulty ratings. From this pool, maintainers curated the final evaluated set used in the main experiments.

\subsection{Contributing Principles}

Contributors were required to satisfy explicit quality requirements designed to prevent shortcuts and ensure that each task measures Skills efficacy rather than prompt-following.

\paragraph{Human-authored instructions.}
Task instructions must be written by humans, not generated by language models. We enforce this because LLM-generated queries can inherit the distributional biases of the same systems under evaluation and often lack the specificity needed for realistic agentic tasks.

\paragraph{Skill generality.}
Skills must provide procedural guidance for a \emph{class} of tasks, not solutions to specific instances. Task instructions must not reference which Skills to use, requiring agents to discover and apply relevant Skills autonomously.

\paragraph{Deterministic verification.}
All success criteria must be testable through programmatic assertions. We target the minimal number of tests needed for verification, avoiding both insufficient coverage and redundant test bloat that can artificially depress pass rates. Tests must include informative error messages and use parametrization rather than duplication when possible.

\subsection{Automated Validation}

Each submission undergoes automated validation before human review:
\begin{itemize}[nosep,leftmargin=*]
    \item \textbf{Structural validation}: Required files must be present (\texttt{task.md} with \texttt{oracle/solve.sh} and \texttt{verifier/test\_outputs.py}; \appautoref{app:task-spec}), and directory layout and frontmatter syntax must be valid.
    \item \textbf{Oracle execution}: The reference solution must achieve a 100\% verifier pass rate. Tasks with failing oracles are rejected.
    \item \textbf{Instruction quality}: Instructions must be human-written, verified through both human review and GPTZero screening. We additionally score instructions on explicit output paths, structured requirements, success criteria, listed constraints, and context-first ordering.
    \item \textbf{Leakage audit}: CI checks for potential Skill-solution leakage, including task-specific constants, filenames, paths, expected outputs, and hard-coded command sequences.
\end{itemize}

\subsection{Human Review}

After automated checks pass, maintainers manually review each task using five criteria:
\begin{enumerate}[nosep,leftmargin=*]
\item \textbf{Data validity}: Input data should reflect real-world complexity; synthetic or toy data is rejected unless explicitly justified.
\item \textbf{Task realism}: Scenarios should reflect realistic professional workflows without artificial difficulty.
\item \textbf{Oracle quality}: Reference solutions should match how domain experts would solve the task.
\item \textbf{Skill quality}: Skills must be error-free, internally consistent, and useful for similar tasks beyond this benchmark.
\item \textbf{Anti-cheating}: Tasks must prevent shortcut solutions such as editing input data, extracting answers from test files, or exploiting verifier implementation details.
\end{enumerate}

\noindent Reviewers run benchmark experiments with and without Skills across multiple agents to confirm that each task provides meaningful signal about Skill efficacy. \autoref{fig:pipeline} provides an end-to-end view of the benchmark construction, quality filtering, and evaluation pipeline.

\subsection{Leakage Prevention}

To prevent Skills from encoding task-specific solutions, we enforce explicit authoring guidelines and conduct leakage audits. A Claude Code Agent SDK-based validation agent runs in CI to detect potential Skill-solution leakage; failed tasks are rejected. Skills must not contain task-specific filenames, paths, identifiers, constants, magic numbers, values from task specifications, exact command sequences that solve benchmark tasks, references to specific test cases, or expected outputs.

Skills must apply to a class of tasks rather than a single instance, provide procedural guidance about how to approach the task class rather than declarative answers about what to output, and be authored independently of benchmark specifications. These rules are central to the paired design: if a Skill contains the answer path for one benchmark instance, the with-Skills condition would no longer measure reusable procedural augmentation.

\subsection{Benchmark Composition}

\begin{table}[htb]
\centering
\small
\caption{Task difficulty stratification based on estimated human completion time in the accepted task pool.}
\label{tab:levels}
\begin{tabular}{lcc}
\toprule
\textbf{Difficulty} & \textbf{Tasks} & \textbf{Human Time} \\
\midrule
Core & 6 (6.9\%) & $<$60 min\\
Extended & 53 (60.9\%) & 1--4 hours\\
Extreme & 28 (32.2\%) & $>$4 hours \\
\bottomrule
\end{tabular}
\end{table}

The current \benchmarkName{} task inventory comprises 87 tasks across 8 domains. Task difficulty is measured by estimated completion time for median specialists in the task domain, without assistance from AI tools. Original contributors provided time estimates, and maintainers with matching domain expertise reviewed those estimates.

\subsection{Domain Taxonomy and Per-Task Mapping}
\label{app:domain-mapping}

We classify each released task into one of eight domains. The taxonomy uses broad top-level labels for professional context while preserving finer distinctions in task metadata. Cybersecurity and Media \& Content Production are currently below $N=8$, so per-domain inferential claims should treat those slices as descriptive until additional tasks are added.

\autoref{tab:task-domain-mapping} lists every task in the current 87-task inventory, its primary capability axis (a separate cut from the goodtask-v2 framework: Reasoning, Agentic Coding, Multimodal, Tool Use, or Search \& Research), and its difficulty marker. The mapping is generated automatically from \texttt{scripts/taxonomy.csv} by \texttt{scripts/distribution.py} so it stays in sync with the figures referenced in \S\ref{sec:SkillsBench}.


{\small
\begin{longtable}{p{0.55\linewidth} p{0.28\linewidth} c}
\caption{Per-domain task listing for the released 87-task \benchmarkName{}. Difficulty marker: \textbf{C}=Core (<60\,min), \textbf{X}=Extended (1--4\,h), \textbf{E}=Extreme ($>$4\,h). Capability is the primary skill the task exercises.}\label{tab:task-domain-mapping}\\
\toprule
\textbf{Task} & \textbf{Capability} & \textbf{Diff.} \\
\midrule
\endfirsthead
\multicolumn{3}{l}{\textit{(continued from previous page)}}\\
\toprule
\textbf{Task} & \textbf{Capability} & \textbf{Diff.} \\
\midrule
\endhead
\bottomrule
\multicolumn{3}{r}{\textit{(continued on next page)}}\\
\endfoot
\bottomrule
\endlastfoot
\multicolumn{3}{p{0.95\linewidth}}{\textbf{Software Engineering} ($n=16$). \textit{Code implementation, debugging, build repair, migration, testing, repo analytics.}} \\
\midrule
\texttt{azure-bgp-oscillation-route-leak} & Reasoning & X \\
\texttt{data-to-d3} & Agentic Coding & X \\
\texttt{debug-trl-grpo} & Agentic Coding & E \\
\texttt{dialogue-parser} & Agentic Coding & C \\
\texttt{fix-build-agentops} & Agentic Coding & C \\
\texttt{fix-build-google-auto} & Agentic Coding & C \\
\texttt{fix-visual-stability} & Agentic Coding & E \\
\texttt{flink-query} & Agentic Coding & E \\
\texttt{jax-computing-basics} & Agentic Coding & X \\
\texttt{llm-prefix-cache-replay} & Agentic Coding & X \\
\texttt{parallel-tfidf-search} & Agentic Coding & X \\
\texttt{python-scala-translation} & Agentic Coding & X \\
\texttt{react-performance-debugging} & Agentic Coding & E \\
\texttt{simpo-code-reproduction} & Agentic Coding & E \\
\texttt{spring-boot-jakarta-migration} & Agentic Coding & E \\
\texttt{tictoc-unnecessary-abort-detection} & Reasoning & E \\
\midrule
\multicolumn{3}{p{0.95\linewidth}}{\textbf{Industrial \& Physical Systems} ($n=14$). \textit{Power systems, manufacturing, robotics/control, construction, and physical simulation.}} \\
\midrule
\texttt{3d-scan-calc} & Reasoning & E \\
\texttt{ada-bathroom-plan-repair} & Reasoning & E \\
\texttt{adaptive-cruise-control} & Reasoning & X \\
\texttt{drone-planning-control} & Reasoning & X \\
\texttt{dynamic-object-aware-egomotion} & Reasoning & X \\
\texttt{energy-ac-optimal-power-flow} & Reasoning & X \\
\texttt{energy-market-pricing} & Reasoning & E \\
\texttt{energy-unit-commitment} & Reasoning & E \\
\texttt{grid-dispatch-operator} & Reasoning & X \\
\texttt{hvac-control} & Reasoning & X \\
\texttt{manufacturing-codebook-normalization} & Reasoning & X \\
\texttt{manufacturing-equipment-maintenance} & Reasoning & X \\
\texttt{manufacturing-fjsp-optimization} & Reasoning & X \\
\texttt{r2r-mpc-control} & Reasoning & X \\
\midrule
\multicolumn{3}{p{0.95\linewidth}}{\textbf{Natural Science} ($n=14$). \textit{Astronomy, seismology, hydrology, materials, physics, and biomedical analysis.}} \\
\midrule
\texttt{crystallographic-wyckoff-position-analysis} & Reasoning & X \\
\texttt{earthquake-phase-association} & Reasoning & E \\
\texttt{earthquake-plate-calculation} & Reasoning & X \\
\texttt{exoplanet-detection-period} & Reasoning & X \\
\texttt{flood-risk-analysis} & Reasoning & X \\
\texttt{glm-lake-mendota} & Reasoning & E \\
\texttt{gravitational-wave-detection} & Reasoning & X \\
\texttt{lab-unit-harmonization} & Reasoning & X \\
\texttt{lake-warming-attribution} & Reasoning & X \\
\texttt{mars-clouds-clustering} & Reasoning & E \\
\texttt{protein-expression-analysis} & Reasoning & X \\
\texttt{quantum-numerical-simulation} & Reasoning & X \\
\texttt{radar-vital-signs} & Reasoning & X \\
\texttt{seismic-phase-picking} & Reasoning & E \\
\midrule
\multicolumn{3}{p{0.95\linewidth}}{\textbf{Office \& White Collar} ($n=14$). \textit{Office documents, spreadsheets, presentations, forms, PDFs, OCR, and enterprise search.}} \\
\midrule
\texttt{citation-check} & Search \& Research & X \\
\texttt{court-form-filling} & Tool Use & C \\
\texttt{edit-pdf} & Tool Use & X \\
\texttt{enterprise-information-search} & Search \& Research & E \\
\texttt{exceltable-in-ppt} & Tool Use & X \\
\texttt{jpg-ocr-stat} & Multimodal & E \\
\texttt{latex-formula-extraction} & Multimodal & X \\
\texttt{offer-letter-generator} & Tool Use & C \\
\texttt{organize-messy-files} & Tool Use & X \\
\texttt{paper-anonymizer} & Multimodal & X \\
\texttt{pdf-excel-diff} & Multimodal & X \\
\texttt{powerlifting-coef-calc} & Tool Use & C \\
\texttt{pptx-reference-formatting} & Tool Use & X \\
\texttt{sales-pivot-analysis} & Tool Use & X \\
\midrule
\multicolumn{3}{p{0.95\linewidth}}{\textbf{Finance \& Economics} ($n=9$). \textit{Financial modeling, accounting, macroeconomic analysis, risk, and fraud workflows.}} \\
\midrule
\texttt{econ-detrending-correlation} & Tool Use & X \\
\texttt{financial-modeling-qa} & Multimodal & E \\
\texttt{invoice-fraud-detection} & Multimodal & E \\
\texttt{reserves-at-risk-calc} & Tool Use & X \\
\texttt{sec-financial-report} & Search \& Research & E \\
\texttt{shock-analysis-demand} & Tool Use & X \\
\texttt{shock-analysis-supply} & Tool Use & E \\
\texttt{weighted-gdp-calc} & Tool Use & X \\
\texttt{xlsx-recover-data} & Tool Use & X \\
\midrule
\multicolumn{3}{p{0.95\linewidth}}{\textbf{Mathematics \& OR} ($n=8$). \textit{Formal proofs, optimization, planning, routing, scheduling, and combinatorial reasoning.}} \\
\midrule
\texttt{bike-rebalance} & Reasoning & X \\
\texttt{civ6-adjacency-optimizer} & Reasoning & E \\
\texttt{exam-block-sequencing} & Reasoning & E \\
\texttt{lean4-proof} & Reasoning & X \\
\texttt{paratransit-routing} & Reasoning & E \\
\texttt{pddl-airport-planning} & Reasoning & X \\
\texttt{pddl-tpp-planning} & Reasoning & X \\
\texttt{travel-planning} & Search \& Research & X \\
\midrule
\multicolumn{3}{p{0.95\linewidth}}{\textbf{Cybersecurity} ($n=7$). \textit{CVE remediation, IDS, fuzzing, dependency audit, network security.}} \\
\midrule
\texttt{dapt-intrusion-detection} & Multimodal & E \\
\texttt{fix-druid-loophole-cve} & Agentic Coding & E \\
\texttt{fix-erlang-ssh-cve} & Agentic Coding & E \\
\texttt{setup-fuzzing-py} & Agentic Coding & X \\
\texttt{software-dependency-audit} & Search \& Research & X \\
\texttt{suricata-custom-exfil} & Multimodal & X \\
\texttt{syzkaller-ppdev-syzlang} & Agentic Coding & X \\
\midrule
\multicolumn{3}{p{0.95\linewidth}}{\textbf{Media \& Content Production} ($n=5$). \textit{Video, audio, 3D content, dubbing, media transformation, and content production.}} \\
\midrule
\texttt{mario-coin-counting} & Multimodal & X \\
\texttt{multilingual-video-dubbing} & Multimodal & X \\
\texttt{threejs-structure-parser} & Multimodal & X \\
\texttt{threejs-to-obj} & Multimodal & X \\
\texttt{video-silence-remover} & Multimodal & E \\
\end{longtable}
}

\clearpage

\vspace{-8pt}

\section{Experimental Protocol Details}
\label{app:experimental-protocol-details}

This appendix expands the concise experimental setup in \S\ref{sec:experimental-setup}. It is separated from the benchmark-construction appendix to keep the evaluation protocol, agent configurations, and scoring choices easy to audit.

\vspace{-8pt}

\subsection{Model--Harness Configurations}

We evaluate four terminal-agent harnesses in the latest aggregate: OpenHands, Claude Code~\citep{anthropic2025claudecode}, Codex CLI~\citep{openai2025codexcli}, and Gemini CLI~\citep{google2025geminicli}. Each model is evaluated with the harness shown in \appautoref{tab:models-full}. All models use temperature 0. The full model identifiers, harness versions, and selected result counts are listed in \appautoref{tab:models-full}.

These commercial harnesses tightly couple model behavior, context construction, tool interfaces, and execution control. We therefore report results at the model--harness level rather than treating models as isolated components. Claude models have been trained with awareness of the Agent Skills specification~\citep{anthropic2025agentskills}, which may affect how they interpret Skill-formatted instructions.

\vspace{-8pt}

\subsection{Skills Conditions}

Each task in the latest aggregate is evaluated under two matched conditions:
\begin{itemize}[nosep,leftmargin=*]
    \item \textbf{No Skills}: The agent receives the task instruction (the \texttt{task.md} body) and task data, but no \texttt{environment/skills/} content is present.
    \item \textbf{Curated Skills}: The complete \texttt{environment/skills/} directory is available, including \texttt{SKILL.md} files, examples, scripts, and references.
\end{itemize}

\noindent A third condition, self-generated Skills, is evaluated on the three dedicated-harness configurations and reported separately (\appautoref{app:self-gen}); it is not part of the two-condition aggregate in \autoref{tab:main-results}.

\vspace{-8pt}

\subsection{Execution Protocol}

For the curated-Skills condition, task Skills are exposed before the task instruction using each harness's native loading mechanism. The task environment, input data, resource limits, and verifier are otherwise identical across conditions. The agent interacts with the containerized environment until it submits a final answer or artifact. The verifier then runs deterministic assertions and writes a pass/fail result.

Selected rows first require a public result file, complete trajectory metadata, and a healthy verifier-scored pass/fail outcome; timeout rows are used only as failure backfill when healthy replacements are unavailable. Stale, rate-limited, or otherwise unscored runs are treated as incomplete coverage and rerun rather than as benchmark outcomes. Container resources, retry policy, and orchestration settings are summarized in \appautoref{app:exp-details}.

\vspace{-8pt} 

\subsection{Scoring}

The primary score is task-macro pass rate, following Terminal-Bench~\citep{merrill2026terminalbenchbenchmarkingagentshard}. For each task $t$ and condition $c$, we average the three selected public trials:
\[
s_{t,c} = \frac{1}{3} \sum_{i=1}^{3} r_{t,c,i}.
\]
where $r_{t,c,i}\in[0,1]$ is the deterministic verifier reward for trial $i$.
We then average task scores over the fixed set of 87 tasks:
\[
\mathrm{PassRate}(c) = \frac{1}{87}\sum_{t=1}^{87} s_{t,c}.
\]
The main no-Skills and curated-Skills conditions use this fixed three-trial denominator per task--model pair. We report absolute improvement, $\Delta = \mathrm{PassRate}(\text{curated}) - \mathrm{PassRate}(\text{no Skills})$, and normalized gain:
\[
g = \frac{\mathrm{PassRate}(\text{curated}) - \mathrm{PassRate}(\text{no Skills})}{1 - \mathrm{PassRate}(\text{no Skills})}.
\]
Normalized gain measures proportional progress toward perfect performance, but it can overstate small absolute gains near the ceiling. We therefore interpret $g$ jointly with absolute pass-rate deltas. The confidence intervals shown in \autoref{fig:skills_results} use the binomial calculation described in \appautoref{app:details}.

\newpage
\end{document}